\documentclass{article} 
\usepackage{arxiv,times}

\usepackage{amsmath,amsfonts,bm}

\def\eqref#1{equation~\ref{#1}}

\def\1{\bm{1}}

\DeclareMathAlphabet{\mathsfit}{\encodingdefault}{\sfdefault}{m}{sl}
\SetMathAlphabet{\mathsfit}{bold}{\encodingdefault}{\sfdefault}{bx}{n}

\usepackage[utf8]{inputenc} 
\usepackage[T1]{fontenc}    
\definecolor{cvprblue}{rgb}{0.21,0.49,0.74}
\definecolor{darkpastelblue}{rgb}{0.47, 0.62, 0.8}
\definecolor{lightpastelblue}{rgb}{0.75, 0.85, 0.95}

\usepackage[hidelinks,colorlinks=true,linkcolor=darkpastelblue,citecolor=darkpastelblue]{hyperref}
\usepackage[capitalize]{cleveref}
\usepackage{url}            
\usepackage{booktabs}       
\usepackage{amsfonts}       
\usepackage{pifont}
\usepackage{nicefrac}       
\usepackage{microtype}      
\usepackage[dvipsnames,table]{xcolor}  
\usepackage{caption}

\usepackage{tikz}           
\usepackage{pgfplots}       
\pgfplotsset{compat=1.18}   
\usetikzlibrary{pgfplots.groupplots}
\usetikzlibrary{positioning}
\usetikzlibrary{backgrounds}
\usetikzlibrary{matrix,shapes.geometric,patterns,decorations.pathmorphing}
\usetikzlibrary{shadows,calc}
\usetikzlibrary{arrows.meta}
\usetikzlibrary{decorations.markings}
\usetikzlibrary{patterns}

\usepackage{amsmath}
\usepackage{multirow}
\usepackage{epigraph}
\usepackage{subcaption}
\usepackage{adjustbox}
\usepackage{nicematrix}
\usepackage{caption}
\usepackage[scaled=0.85]{beramono}

\usepackage{siunitx}

\newcommand{\greyrule}{\arrayrulecolor{black!30}\midrule\arrayrulecolor{black}}

\newcommand{\mname}{\texttt{DF}}

\colorlet{dfcolor}{RoyalPurple}
\usepackage{fvextra}
\usepackage[most]{tcolorbox}

\newcounter{dfalgo}
\newcommand{\dfalgotitle}[2]{\refstepcounter{dfalgo}\label{#2}\textbf{Algorithm~\thedfalgo}\enspace #1}
\crefformat{dfalgo}{#2Algorithm-#1#3}
\Crefformat{dfalgo}{#2Algorithm-#1#3}
\crefmultiformat{dfalgo}{#2Algorithm-#1#3}{ and #2Algorithm-#1#3}{, #2Algorithm-#1#3}{ and #2Algorithm-#1#3}
\Crefmultiformat{dfalgo}{#2Algorithm-#1#3}{ and #2Algorithm-#1#3}{, #2Algorithm-#1#3}{ and #2Algorithm-#1#3}
\crefrangeformat{dfalgo}{#3Algorithm-#1#4 to #5Algorithm-#2#6}
\Crefrangeformat{dfalgo}{#3Algorithm-#1#4 to #5Algorithm-#2#6}

\colorlet{dfkw}{dfcolor!90!black}
\colorlet{dfcm}{black!50}
\colorlet{dfzero}{black!11}            
\colorlet{dflookahead}{YellowGreen!22} 
\colorlet{dfselfcond}{dfcolor!12}      

\newlength{\dfcodeskip}
\newcommand{\dfcode}[3]{{\setlength{\topsep}{0pt}\setlength{\partopsep}{0pt}%
  \VerbatimInput[commandchars=\\\{\}, fontsize=\fontsize{7.6}{9.6}\selectfont,
                 firstline=#2, lastline=#3]{#1}}}

\newtcolorbox{dfseg}[1]{enhanced, colback=#1, frame hidden, arc=2pt, boxrule=0pt,
  left=0pt, right=0pt, top=0pt, bottom=0pt, boxsep=1.2pt,
  before skip=0pt, after skip=0pt, enlarge left by=-1.2pt, enlarge right by=-1.2pt}

\tcbset{dfalgobox/.style={enhanced, arc=3pt, boxrule=0.6pt, colframe=black!25,
  colback=white, colbacktitle=black!6, coltitle=black, fonttitle=\normalfont,
  titlerule=0.4pt, left=6pt, right=6pt, top=5pt, bottom=5pt, boxsep=0pt,
  toptitle=2.5pt, bottomtitle=2.5pt,
  shadow={0.5ex}{-0.5ex}{0pt}{fill=black, opacity=0.12}}}

\newcommand{\dfswatch}[1]{\tikz\filldraw[fill=#1, draw=black!25, line width=0.3pt, rounded corners=1pt] (0,0) rectangle (0.9em,0.9em);}
\newcommand{\df}[1]{\textbf{#1}}

\definecolor{latentbg}{RGB}{212,220,247}  
\definecolor{pixelbg}{RGB}{253,237,196}   

\newcount\epi
\newcommand{\eppip}[1]{\textcolor{#1}{\rule{0.72ex}{0.72ex}}}
\newcommand{\eprow}[3]{
  \epi=#1\relax
  \loop\ifnum\epi<\numexpr#1+8\relax
    \ifnum\epi<#2\relax\eppip{#3}\else\eppip{gray!13}\fi
    \ifnum\epi<\numexpr#1+7\relax\hspace{0.25ex}\fi
    \advance\epi by 1\relax
  \repeat}
\newcommand{\epbarcore}[3]{
  \makebox[1.9em][r]{#1}\hspace{0.5em}%
  \raisebox{-0.3ex}{\vbox{\offinterlineskip
    \hbox{\eprow{0}{\numexpr#2/100\relax}{#3}}\kern0.25ex
    \hbox{\eprow{8}{\numexpr#2/100\relax}{#3}}}}}

\newcommand{\abbrcite}[2]{\texttt{#1}: \citealp{#2}}

\title{Dino Forcing Flow Models: Do not denoise\\ what you can predict}

\author{Arijit Ghosh $^{*,1,2}$, Lucas Degeorge $^{*,1,2,4}$, Paul Couairon $^{2}$ \\
\textbf{Alexei A. Efros $^{3}$, Vicky Kalogeiton $^{\dagger,2}$, David Picard $^{\dagger,1}$}  \\ \\
$^*$ Equal Contribution, $^\dagger$ Equal Supervision\\
$^1$ ENPC, IP Paris, $^2$ Ecole Polytechnique, IP Paris, $^3$ UC Berkeley, $^4$ AMIAD
}

\newsavebox{\plotbox}\newsavebox{\gridbox}
\newlength{\Hp}\newlength{\Hg}\newlength{\leftW}\newlength{\cw}
\newlength{\gsep}\newlength{\plotW}\newlength{\panelgap}

\newlength{\stripeht}\newlength{\stripedp}
\newcommand{\stripe}[1]{%
  \setlength{\stripeht}{\arraystretch\ht\strutbox}%
  \setlength{\stripedp}{\arraystretch\dp\strutbox}%
  \tikz[overlay]\fill[#1] (0,-\stripedp) rectangle (0.6mm,\stripeht);}
\iclrfinalcopy 
\begin{document}

\maketitle

\begin{abstract}
Co-denoising pretrained representations such as DINO can substantially improve the training speed and quality of flow matching models, but it introduces a second denoising trajectory and requires carefully designed schedules. We propose a simpler alternative: predict the pretrained representation directly, then condition the model on its own prediction. This removes the need for a second ODE and any representation-specific denoising schedules, while retaining the benefits of representation guidance. Our approach converges substantially faster and achieves better generation quality as measured by FID score. On ImageNet unguided setting, it outperforms the state of the art in latent space at  8$\times$ fewer epochs than prior methods; in pixel space, it improves over comparable prior methods by more than 20\%. These results support a simple principle: do not denoise what you can predict. Our code is openly available in \href{https://github.com/arijit-hub/dino_forcing}{\texttt{https://github.com/arijit-hub/dino\_forcing}}. 
\end{abstract}

\begin{figure*}[htbp]
    \input{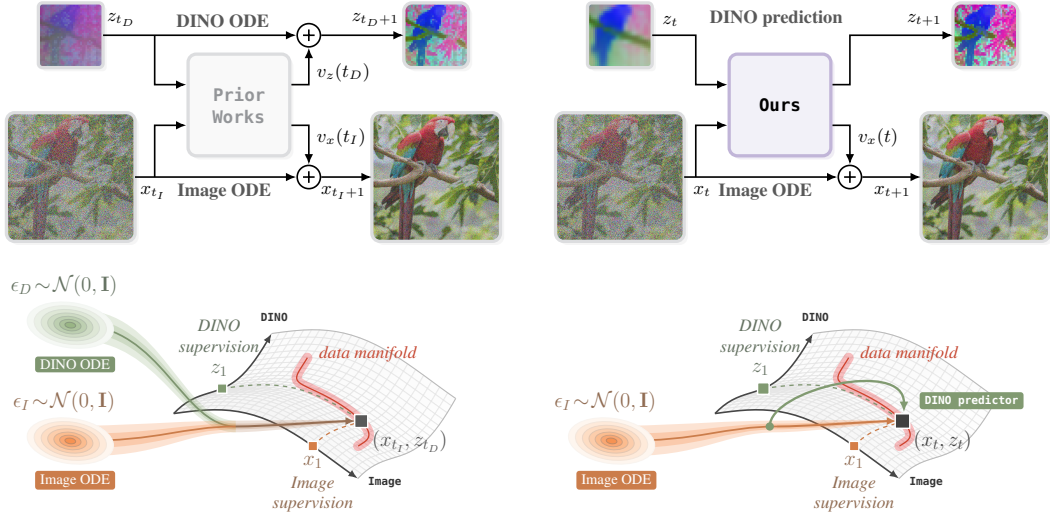}
    \caption{%
    \textbf{Left:} Prior co-denoising methods generate both the image and the pretrained representation from noise by jointly solving their individual ODEs through iterative modification during sampling. The model predicts the velocity for both modalities at different time schedules (\textit{top}) which leads to 2 trajectories starting from independent noise samples and arriving at the same pair (\textit{bottom}).
    \textbf{Right:} Our proposed \texttt{\textbf{Dino Forcing}} (\texttt{\textbf{DF}}) method keeps the ODE only for the image generation since the representation is deterministically tied to the image. The model predicts the target representation in addition to the image velocity (\textit{top}), providing a self-condition to guide the image ODE (\textit{bottom}). 
    }
    \label{fig:teaser}
\end{figure*}
 
\section{Introduction }
\label{sec:intro}
\begin{figure*}[t]
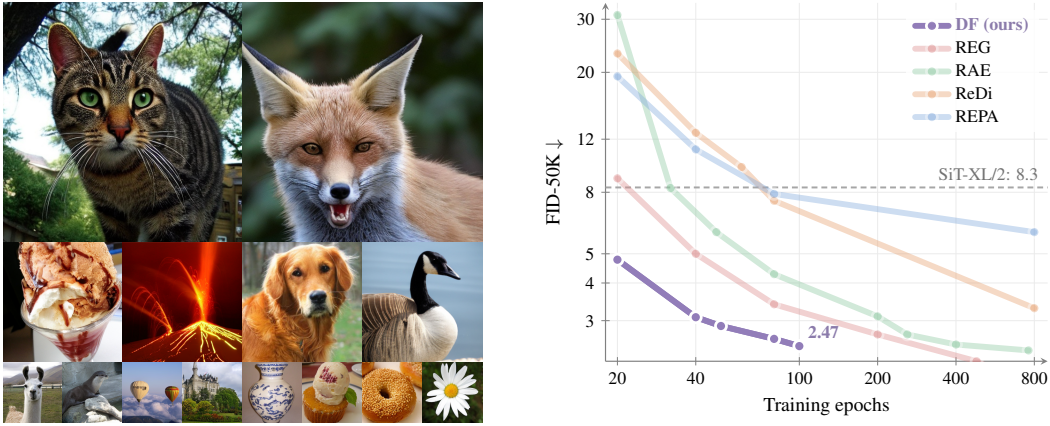

    \centering
    \setlength{\fboxsep}{0pt}
    \setlength{\plotW}{0.50\textwidth}    
    \setlength{\gsep}{0pt}              
    \setlength{\panelgap}{0.02\textwidth} 
    %
    \savebox{\plotbox}{\resizebox{\plotW}{!}{\begingroup
\renewcommand\familydefault{\sfdefault}

\definecolor{cReDi}  {RGB}{232,160,110}   
\definecolor{cRAE}   {RGB}{140,200,160}   
\definecolor{cREPA}  {RGB}{130,170,220}
\definecolor{cREGLUE}{RGB}{235,160,190}   
\definecolor{cREG}   {RGB}{220,140,140}   
\colorlet{dfcolor}{RoyalPurple!60}

\begin{tikzpicture}
\pgfplotsset{
  curve/.style={
    line width=3pt, line join=round, line cap=round,
    mark=*, mark size=2.8pt,
    mark options={solid, draw=white, line width=1.4pt},
  },
}

\begin{axis}[
  width=9.5cm, height=8cm,
  xmode=log, ymode=log,
  xlabel={Training epochs},
  ylabel={FID-50K $\downarrow$},
  label style={font=\normalsize},
  tick label style={font=\small},
  xmin=18, xmax=900,
  ymin=1.6, ymax=34,
  xtick={20,40,100,200,400,800},
  xticklabels={20,40,100,200,400,800},
  ytick={2,3,4,5,8,12,20,30},
  yticklabels={2,3,4,5,8,12,20,30},
  minor tick num=0,
  log ticks with fixed point,
  grid=major,
  major grid style={draw=black!8, line width=0.4pt},
  axis line style={draw=black!50, line width=0.6pt},
  tick style={draw=black!50},
  axis lines=left,
  enlargelimits=false,
    clip=true,                 
  clip mode=individual,      
  clip marker paths=true,    
  set layers,          
  legend cell align=left,
  legend style={
    at={(0.98,0.98)}, anchor=north east,
    draw=none, fill=white, fill opacity=0.85, text opacity=1,
    font=\small, row sep=0pt,
  },
  legend image code/.code={\draw[mark repeat=2, mark phase=2, #1]
    plot coordinates {(0cm,0cm) (0.3cm,0cm) (0.6cm,0cm)};},
     baseline/.style={curve, opacity=0.4},
]

\addplot[draw=black!35, densely dashed, line width=1pt, forget plot]
  coordinates {(18,8.3) (900,8.3)};
\node[anchor=south east, font=\footnotesize, text=black!55]
  at (axis cs:880,8.3) {SiT-XL/2: 8.3};

\addplot[curve, color=dfcolor] coordinates {
  (20,3.95) (40,2.45) (50,2.24) (80,1.93) (100,1.84)
};
\addlegendentry{\textbf{\textcolor{dfcolor!85!black}{DF (ours)}}}

\addplot[baseline, color=cREG] coordinates {
  (20,8.9) (40,5.0) (80,3.4) (200,2.7) (480,2.2) (800,1.8)
};
\addlegendentry{REG}

\addplot[baseline, color=cRAE] coordinates {
  (20,31) (32,8.26) (48,5.90) (80,4.28) (200,3.1) (260,2.7) (400,2.5) (760,2.39)
};
\addlegendentry{RAE}

\addplot[baseline, color=cReDi] coordinates {
  (20,23.1) (40,12.6) (60,9.7) (80,7.5) (800,3.3)
};
\addlegendentry{ReDi}

\addplot[baseline, color=cREPA] coordinates {
  (20,19.4) (40,11.1) (80,7.9) (800,5.9)
};
\addlegendentry{REPA}

\addplot[curve, color=dfcolor, forget plot] coordinates {
  (20,3.95) (40,2.45) (50,2.24) (80,1.93) (100,1.84)
};

\node[anchor=south west, font=\small\bfseries, text=dfcolor!85!black,
      inner sep=1pt] at (axis cs:106,1.91) {1.84};

\end{axis}
\end{tikzpicture}
\endgroup}}%
    \setlength{\Hp}{\dimexpr\ht\plotbox+\dp\plotbox\relax}%
    %
    \setlength{\leftW}{\dimexpr\Hp*8/7\relax}%
    %
    \savebox{\gridbox}{\vbox{\offinterlineskip
        \setlength{\cw}{\dimexpr(\leftW-\gsep)/2-1sp\relax}%
        \hbox to\leftW{\hss\foreach \n in {1,2}{\includegraphics[width=\cw,height=\cw]{images/raw_jpg_pictures/qualitative_samples/r1_\n}\ifnum\n<2\relax\hspace{\gsep}\fi}\hss}%
        \vskip\gsep
        \setlength{\cw}{\dimexpr(\leftW-3\gsep)/4-1sp\relax}%
        \hbox to\leftW{\hss\foreach \n in {1,...,4}{\includegraphics[width=\cw,height=\cw]{images/raw_jpg_pictures/qualitative_samples/r2_\n}\ifnum\n<4\relax\hspace{\gsep}\fi}\hss}%
        \vskip\gsep
        \setlength{\cw}{\dimexpr(\leftW-7\gsep)/8-1sp\relax}%
        \hbox to\leftW{\hss\foreach \n in {1,...,8}{\includegraphics[width=\cw,height=\cw]{images/raw_jpg_pictures/qualitative_samples/r3_\n}\ifnum\n<8\relax\hspace{\gsep}\fi}\hss}%
    }}%
    %
    \setlength{\Hg}{\dimexpr\ht\gridbox+\dp\gridbox\relax}%
    \resizebox{\linewidth}{!}{%
      $\vcenter{\hbox{\usebox{\gridbox}}}$%
      \hspace{\panelgap}%
      $\vcenter{\hbox{\resizebox{!}{\Hg}{\begingroup
\renewcommand\familydefault{\sfdefault}

\definecolor{cReDi}  {RGB}{232,160,110}   
\definecolor{cRAE}   {RGB}{140,200,160}   
\definecolor{cREPA}  {RGB}{130,170,220}
\definecolor{cREGLUE}{RGB}{235,160,190}   
\definecolor{cREG}   {RGB}{220,140,140}   
\colorlet{dfcolor}{RoyalPurple!60}

\begin{tikzpicture}
\pgfplotsset{
  curve/.style={
    line width=3pt, line join=round, line cap=round,
    mark=*, mark size=2.8pt,
    mark options={solid, draw=white, line width=1.4pt},
  },
}

\begin{axis}[
  width=9.5cm, height=8cm,
  xmode=log, ymode=log,
  xlabel={Training epochs},
  ylabel={FID-50K $\downarrow$},
  label style={font=\normalsize},
  tick label style={font=\small},
  xmin=18, xmax=900,
  ymin=1.6, ymax=34,
  xtick={20,40,100,200,400,800},
  xticklabels={20,40,100,200,400,800},
  ytick={2,3,4,5,8,12,20,30},
  yticklabels={2,3,4,5,8,12,20,30},
  minor tick num=0,
  log ticks with fixed point,
  grid=major,
  major grid style={draw=black!8, line width=0.4pt},
  axis line style={draw=black!50, line width=0.6pt},
  tick style={draw=black!50},
  axis lines=left,
  enlargelimits=false,
    clip=true,                 
  clip mode=individual,      
  clip marker paths=true,    
  set layers,          
  legend cell align=left,
  legend style={
    at={(0.98,0.98)}, anchor=north east,
    draw=none, fill=white, fill opacity=0.85, text opacity=1,
    font=\small, row sep=0pt,
  },
  legend image code/.code={\draw[mark repeat=2, mark phase=2, #1]
    plot coordinates {(0cm,0cm) (0.3cm,0cm) (0.6cm,0cm)};},
     baseline/.style={curve, opacity=0.4},
]

\addplot[draw=black!35, densely dashed, line width=1pt, forget plot]
  coordinates {(18,8.3) (900,8.3)};
\node[anchor=south east, font=\footnotesize, text=black!55]
  at (axis cs:880,8.3) {SiT-XL/2: 8.3};

\addplot[curve, color=dfcolor] coordinates {
  (20,3.95) (40,2.45) (50,2.24) (80,1.93) (100,1.84)
};
\addlegendentry{\textbf{\textcolor{dfcolor!85!black}{DF (ours)}}}

\addplot[baseline, color=cREG] coordinates {
  (20,8.9) (40,5.0) (80,3.4) (200,2.7) (480,2.2) (800,1.8)
};
\addlegendentry{REG}

\addplot[baseline, color=cRAE] coordinates {
  (20,31) (32,8.26) (48,5.90) (80,4.28) (200,3.1) (260,2.7) (400,2.5) (760,2.39)
};
\addlegendentry{RAE}

\addplot[baseline, color=cReDi] coordinates {
  (20,23.1) (40,12.6) (60,9.7) (80,7.5) (800,3.3)
};
\addlegendentry{ReDi}

\addplot[baseline, color=cREPA] coordinates {
  (20,19.4) (40,11.1) (80,7.9) (800,5.9)
};
\addlegendentry{REPA}

\addplot[curve, color=dfcolor, forget plot] coordinates {
  (20,3.95) (40,2.45) (50,2.24) (80,1.93) (100,1.84)
};

\node[anchor=south west, font=\small\bfseries, text=dfcolor!85!black,
      inner sep=1pt] at (axis cs:106,1.91) {1.84};

\end{axis}
\end{tikzpicture}
\endgroup}}}$%
    }%
    \caption{\textbf{Left:} Qualitative samples. More examples are given in Figures \ref{fig:qual_sit256_nocfg} to \ref{fig:qual_jit256_cfg}. \textbf{Right:} Convergence speed-up (FID-50K vs.\ training epochs)}.
    \label{fig:convergence}
\end{figure*}

Pretrained representations should help a model generate better images, not strain it by learning one more diffusion trajectory! Yet recent flow matching methods that leverage pretrained representations, such as \texttt{DINO}, typically diffuse and denoise the representation alongside the image. In fact, a trend started by~\citep{bao2023one, xu2023versatile} and recently popularized by ~\citep{kouzelis2026boosting,baade2026latent} is to jointly denoise images and \texttt{DINO} embeddings, followed by ~\citep{wu2026representation, pan2026semantics, petsangourakis2025reglue, lin2026v} that extend this with \texttt{<cls>} token, dual-stream or auxiliary networks. This joint co-denoising speeds up training compared to the standard denoising as the model can leverage the additional information of the frozen representations. 

However, in this work, we claim that this co-denoising comes with two paradoxes: First, the extra information introduced to guide the image generation problem is now completely hidden in the noise such that the model has to untangle the informative bits of that modality. Second, the learning problem is made artificially harder than it should be\footnote{It is still easier than not using the additional representation.} because it consists of matching pairs of independent noise samples ($\epsilon_{I},\epsilon_D$ from \cref{fig:teaser}, bottom left) to a coupled pair (image $x$, DINO $z=\phi(x)$) that does not require the extra noise since the link between the image and its representation is deterministic.

In this work, we remove these paradoxes with one simple idea: pretrained representations should constrain generation, not become
another generation target. The pretrained representation is not an independent quantity that needs its own generative trajectory; it is a property of the image that the model can progressively infer as the image emerges. Our key idea is therefore simple: rather than denoising a second representation, we ask the model to {\em predict} the representation it needs (\cref{fig:teaser} top right).
This keeps the search entirely in the image space, recovering the simple one-to-one mapping problem (\cref{fig:teaser} bottom right) while benefiting from the guidance of the pretrained representation.

This principle leads directly to our proposed \texttt{DINO Forcing} (\texttt{DF}), a method with two ouputs, one from image ODE and one for predicting \texttt{DINO} features that are used as self-conditioning.  During training, a first look-ahead pass predicts the \texttt{DINO} representation associated with the current noisy image; a second pass conditions on this prediction to learn the image velocity. The predicted representation is aligned with a frozen \texttt{DINO} encoder (\cref{fig:arch}, right). At inference, the same prediction is simply rolled forward from one solver step to the next (\cref{fig:sampling}). Thus, the pretrained representation is produced by the generator itself and evolves together with the image.

In latent space, \texttt{DF} reaches an FID of 1.84 after only 100 epochs, surpassing prior methods trained for substantially longer (\cref{fig:convergence}, right). At the same training budget, our \texttt{B} model outperforms \texttt{XL} baseline models, showing that the gain does not come from scale. 
In pixel space, \texttt{DF} improves over \texttt{Latent Forcing}~\citep{baade2026latent} from 7.20 to 5.53 FID at comparable model size and training budget. These results support our claim that explicitly co-denoising is unnecessary: predicting the pretrained representation is not only simpler, but leads to substantially faster convergence.

Our contrbutions are: 
(1) we show that directly predicting \texttt{DINO} representations is sufficiently informative to guide generation, without needing to explicitly denoise it.
(2) we propose \texttt{DINO Forcing}: a schedule-free self-conditioning mechanism that predicts, rather than denoise, the pretrained representation used to guide generation. 
(3) \texttt{DF} achieves strong performance in substantially fewer training epochs for both latent and pixel space flow matching frameworks.
\section{Related Work}
\label{sec:related}

\paragraph{Representation alignment in image generation.}

Diffusion~\citep{sohl2015deep,ho2020denoising} and flow matching~\citep{lipman2023flow} models are trained with denoising objectives that favour fine-grained, reconstruction-oriented representations over high-level semantic ones~\citep{yu2025representation}, which slows convergence: state-of-the-art generators typically need over 500 epochs~\citetext{\abbrcite{DiT}{peebles2023scalable}; \abbrcite{SiT}{ma2024sit}; \abbrcite{JiT}{li2026back}; \abbrcite{fv-loss}{degeorge2026balancing}; \abbrcite{TREAD}{krause2025tread}}. Aligning internal features with external pretrained representations accelerates training~\citetext{\abbrcite{REPA}{yu2025representation}; \abbrcite{iREPA}{singh2026what}; \abbrcite{HASTE}{wang2025repa}; \abbrcite{REED}{wang2026learning}; \abbrcite{DDT}{wang2026ddt}; \abbrcite{PixelDiT}{yu2026pixeldit}; \abbrcite{PixelREPA}{shin2026representation}; \abbrcite{DeCo}{ma2026deco}; \abbrcite{PixelGen}{ma2026pixelgen}}. But at sampling time the model runs unguided: the signal lacks when it is most needed. A complementary line performs flow matching directly in a pre-aligned latent space~\citetext{\abbrcite{RAE}{zheng2026diffusion}; \abbrcite{DINO-SAE}{chang2026dino}}, so the representation is present throughout sampling. But this requires training a new decoder and architectural changes. A more comprehensive comparison is given in Appendix~\ref{app:sec:related_work}.

\paragraph{Co-denoising for image generation.}

Jointly denoising multiple modalities~\citep{bao2023one, xu2023versatile} offers an appealing alternative: it implicitly aligns pretrained representations while preserving them throughout generation, including at inference. This accelerates training convergence, both in a latent~\citetext{\abbrcite{ReDi}{kouzelis2026boosting}; \abbrcite{CoReDi}{kouzelis2026coevolving}; \abbrcite{REG}{wu2026representation}; \abbrcite{SFD}{pan2026semantics}; \abbrcite{REGLUE}{petsangourakis2025reglue}} and directly in pixel space~\citetext{\abbrcite{LatentForcing}{baade2026latent}; \abbrcite{V-Co}{lin2026v}; \abbrcite{CoReDi}{kouzelis2026coevolving}}. The cost is a second noise scheduler that must be hand-crafted: \texttt{SFD}~\citep{pan2026semantics} and \texttt{LatentForcing}~\citep{baade2026latent} both find that denoising the representation to a low noise level \emph{before} starting the image yields the fastest convergence, a schedule that likely needs re-tuning for different modalities.
A more direct approach is to condition the generator on pretrained representations rather than denoise them from scratch~\citetext{\abbrcite{RCG}{li2024return}; \abbrcite{PixelDiT2}{yu2026pixeldit2representationgroundedpixeldiffusion}; \abbrcite{WiT}{wang2026wit}}. Inspired by self-conditioning~\citep{jabri2023scalable, chen2023analog}, we condition the model on its \emph{own predicted representation} at each denoising step. Unlike prior self-conditioning work, we align these predictions with an external pretrained feature space, i.e. \texttt{DINO}. 
The representation is then available throughout generation with no handcrafted schedule, while still training faster. 
\section{DINO-Forcing}
\label{sec:method}

Here, we first give an overview of \mname, then detail its two-pass training procedure (\cref{subsec:self_cond}), and how image and representation tokens are fused, and finally its sampling (\cref{subsec:sampling}).

\paragraph{Overview}
\texttt{DINO-Forcing} (\mname) generates images by relying on a simple idea: rather than denoising a separate pretrained representation in parallel, it \emph{predicts} the pretrained representation it needs (i.e. \texttt{DINO}) and conditions itself on that prediction throughout generation. 

\begin{figure*}[t]
    \centering
    \resizebox{\textwidth}{!}{\input{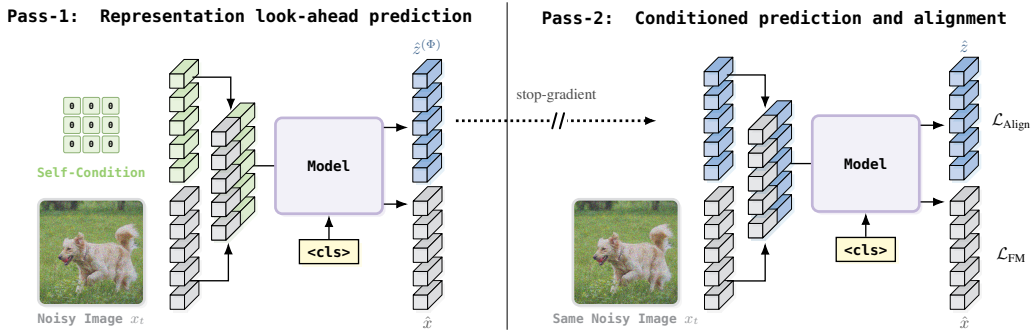}}
    \caption{\textbf{Overview of our proposed training framework}. A first forward pass produces a representation look-ahead (no gradient), which is then used to condition a second pass that optimizes both the flow matching and representation alignment objectives.}
    \label{fig:arch}
\end{figure*}

Concretely, at each training step, \mname\ produces a \textit{look-ahead} estimate of the pretrained representation, $\hat{z}$, associated with the image, $x$, being denoised (see \cref{fig:arch}; Pass-1). Then, it uses that estimate as a conditioning signal for a forward-backward update pass (see \cref{fig:arch}; Pass-2). At sampling time, these \texttt{DINO} representations are predicted and are rolled out across denoising steps, providing a continuously updated, model-generated guide (see \cref{fig:sampling} and \cref{alg:sample}).

\begin{figure*}[t]
    \centering
    \input{algorithms/pseudocode}
\end{figure*}

\subsection{Self-Conditioning on Predicted Representations}  
\label{subsec:self_cond}

 The core idea of our proposed \mname\ is a two-pass training procedure, illustrated in \cref{fig:arch} and detailed in \cref{alg:train}. 
 For the remainder of this section, for simplicity, we omit referring to additional conditions, such as class or text; however, in our experiments we do include them.

\paragraph{Pass 1: Representation look-ahead prediction.}
We perform a forward pass with $x_t$ as the image input and a zeros tensor, $\mathbf{0}$, as placeholder for the representation conditioning. This produces a preliminary representation prediction, $\hat{z}^{(\Phi)}$, which serves as a look-ahead estimate of the content in the current denoising trajectory (similar to~\cite{jabri2023scalable, chen2023analog, dufour2024don}; left part of \cref{fig:arch}). This pass runs without gradients: it provides the conditioning for Pass 2.

\paragraph{Pass 2: Conditioned prediction and alignment.} 
The look-ahead prediction $\hat{z}^{(\Phi)}$ is then used as the conditioning input in a second forward pass. This pass produces two outputs: (i) the flow matching velocity prediction $v_\theta(x_t| \hat{z}^{(\Phi)}, t)$ for image generation, and (ii) a refined representation prediction $\hat{z}$ that is supervised against an external pretrained feature space, $\phi(x)$ (contrary to prior methods ~\cite{jabri2023scalable, chen2023analog, dufour2024don} who leave this added condition unsupervised). As such, the final training objective is formulated as:
\begin{equation}
    \mathcal{L}_{\text{total}} = \mathcal{L}_{\text{FM}} + \lambda \cdot \mathcal{L}_{\text{Align}},
    \label{eq:loss}
\end{equation}
with $\lambda$ a scalar balancing the two objectives: $\mathcal{L}_{\text{FM}} = \|v_\theta - (x - \epsilon)\|^2$ is the standard flow matching velocity loss~\citep{lipman2023flow}, and $\mathcal{L}_{\text{Align}}$ is the loss between the predicted and frozen representation. In this work, we follow \texttt{REPA} formulation~\citep{yu2025representation} and define $\mathcal{L}_{\text{Align}}$ as a cosine similarity loss between $\hat{z}$ and target representations $\phi(x)$ extracted from the clean image $x$. This is given as follows:
\begin{equation}
\mathcal{L}_{\text{Align}}=-\left\langle\frac{\hat{z}}{\|\hat{z}\|}, \frac{\phi(x)}{\|\phi(x)\|}\right\rangle 
\end{equation}

In this work, $\phi(x)$ denotes \texttt{DINO} features, following~\citep{kouzelis2026boosting, baade2026latent}. Note that the formulation is not tied to this choice: any other comparative loss could be used instead.

With this two-pass design, the model learns two things jointly: generating images and predicting accurate representations. These predictions are fed back as conditioning in the same training step.

\paragraph{Fusing Image and Frozen Representations. }
\mname\ takes as input merged image and pretrained representation (i.e.\ \texttt{DINO}) tokens. For merging, we concatenate them along the channel dimension (see \cref{fig:arch}), keeping the token count identical to the standard image-only model, which makes our work compatible with existing flow models. More details in Appendix~\ref{app:sub:fuse}, and a discussion of other fusion strategies in \citet{lin2026v}.

\subsection{Sampling}
\label{subsec:sampling}

\begin{figure*}[t]
    \centering
    \resizebox{\textwidth}{!}{\input{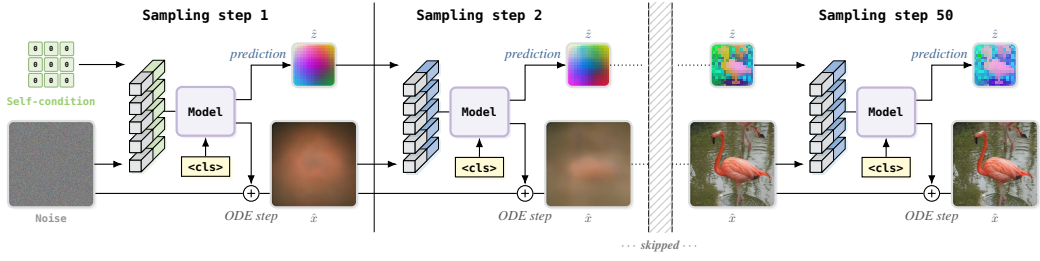}}
    \caption{\textbf{Overview of our sampling framework.} At each step the model outputs a \emph{ODE step}, and a \emph{prediction} $\hat z$ used as the self-condition at the next step. Note that $\hat z$ is not a denoised feature map but the model's prediction of the DINO features of the denoised image $\hat x$.}
    \label{fig:sampling}
\end{figure*}

At inference, we replicate the structure of training through a representation roll-out across denoising steps, illustrated in \cref{fig:sampling} (similar to ~\cite{jabri2023scalable,dufour2024don}).

We initialize the image input with pure noise $x_0 \sim \mathcal{N}(0, I)$ and set the starting representation to a zero tensor $\mathbf{0}$, mirroring Pass~1 at training time (see \cref{subsec:self_cond}). At each denoising step $t$, the model produces both a velocity estimate, used to update $x_{t}$ via the ODE solver, and a representation prediction $\hat{z}$. This prediction is then passed as the conditioning input for the \emph{next} denoising step, replacing the zeros placeholder. The roll-out continues for all steps of the solver.

\begin{figure*}[h]
    \centering
    \input{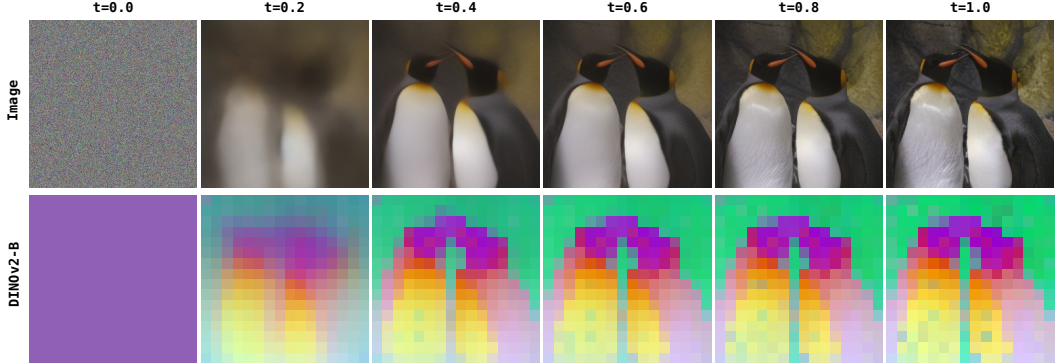}
    \caption{Evolution of the image and the PCA of the predicted \texttt{DINO} features throughout sampling.}
    \label{fig:codenoising}
\end{figure*}

The result is a sampling procedure in which the representation conditioning signal grows progressively more coherent as the image takes shape (see \cref{fig:codenoising} and \cref{fig:qual_dino_progression}), providing structured guidance precisely when it is most useful, without any modification to the standard denoising schedule.

\section{Prediction vs co-denoising}
\label{sec:co-denoising}
As motivated in \cref{sec:related}, most recent image generation approaches either remove pretrained representation information at inference time, or require extra decoders and hand-crafted denoising schedules to preserve it. 

We focus on co-denoising, as it keeps an estimate of the pretrained representation during the denoising process, and try to directly address the limitations of these approaches. 
Co-denoising methods~\citep{bao2023one,kouzelis2026boosting,baade2026latent} come with notable performance gains, mainly due to sharing one network between two tasks: denoising the image $x$ and denoising an additional pretrained representation $\phi(x)$, which can be text as in \citep{bao2023one} or \texttt{DINOv2} as in \citep{kouzelis2026boosting, baade2026latent}. Even though this results in faster convergence than image-only denoising baselines, it requires an additional ODE trajectory for the pretrained representations (and thereby handcrafted schedules). 

Specifically, with independent noises $\epsilon_I,\epsilon_D$ and two time variables, the co-denoising model fits
$$v_\theta\big(x_{t_I},\,\phi(x)_{t_D},\,t_I,\,t_D\big), \qquad (t_I,t_D)\in[0,1]^2 \quad ,$$
so it must be accurate over the whole square of noise-level pairs (\cref{fig:teaser}, bottom left).

Instead, we claim that this additional trajectory can be omitted. Since $\phi$ is deterministic, the pair $(x,\phi(x))$ never fills the product space, but it lies on
$$\{(x,\phi(x)) : x \in \mathcal{X}\} \subset \mathbb{R}^{d_I}\times\mathbb{R}^{d_D} \quad ,$$
and the pretrained representation coordinate adds no degrees of freedom to the distribution being modeled: 
given a clean image there is no distribution over $\phi(x)$ to sample from. Given a noisy $x_t$, the best estimate of the pretrained representation is $\hat{z} = \mathbb{E}[\phi(x)\mid x_t]$, which depends on $x_t$ only with no noise level of its own. A forward pass of the network is enough to predict it, and needs no schedule. As such, in \texttt{DINO Forcing} we ask the model to simply \emph{predict} the pretrained representation it needs and condition itself on that prediction throughout generation (\cref{fig:sampling}).

This results in us keeping the shared network and its two tasks, but it removes the quadratic search of two ODE trajectories (the time domain collapses from $[0,1]^2$ back to $[0,1]$) (\cref{fig:teaser}, right) and the corresponding handcrafted schedules.
\section{Experiments}
\label{sec:experiments}

We evaluate our method on the ImageNet class-conditional benchmark, in both latent space (\cref{sec:latentspace}) and pixel space (\cref{sec:pixelspace}). We then analyse what the representation prediction branch learns by inverting its predictions back to images (\cref{sec:dino_inversion}), and study generation driven by the predicted features alone (\cref{sec:dino_cond_only}). We ablate the components of our method in \cref{sec:ablation}. Qualitative samples are shown in \cref{fig:convergence} left, with more uncurated samples in Appendix \ref{app:sec:qualitative_results}.

\paragraph{Evaluation Metrics} We compare \df{\texttt{DF}} against latent space state-of-the-art methods using the \texttt{ADM}~\citep{dhariwal2021diffusion} evaluation suite and its reference statistics. We generate 50K samples and report Fréchet Inception Distance (\textbf{FID})~\citep{heusel2017gans}, Inception Score (\textbf{IS})~\citep{salimans2016improved}, Precision (\textbf{P}) and Recall (\textbf{R})~\citep{kynkaanniemi2019improved}. For pixel space, we report \textbf{FID} against the \texttt{JiT} reference statistics. Since \texttt{JiT} does not release the images needed for the remaining metrics, \textbf{IS}, \textbf{P} and \textbf{R} are computed with the \texttt{ADM} suite.

\subsection{Latent Space Setting}
\label{sec:latentspace}

\paragraph{Setup} In the latent space setting, we follow the implementations of \texttt{SiT}~\citep{ma2024sit} with the only exception of having extra input and output projections for the pretrained representation features. We train our model in the \texttt{sd-vae-ft-ema}~\citep{rombach2022high} latent space. We follow prior works~\citep{yu2025representation, baade2026latent, kouzelis2026boosting} and use \texttt{DINOv2-B}~\citep{oquab2024dinov} features. We weight our frozen representation loss ($\lambda$ in \cref{eq:loss}) with a value of $0.5$, similar to prior work~\citep{yu2025representation}. Consistent with previous works~\citep{ma2024sit}, we use the Adam~\citep{kingma2014adam} optimizer without any weight decay and a constant learning rate of $1e{-4}$. In contrast to previous methods, we sample using a simple Euler ODE solver in 50 steps instead of using SDE solvers. We train using the null condition 10\% of the time for classifier-free guidance~\citep{ho2021classifierfree}. For the pretrained representation condition, 10\% of the time we do optimization on just the 1st forward pass to make the model learn a better prediction when the conditions are all zeros. We also set the pretrained representation prediction condition to the original \texttt{DINOv2} features 5\% of the time to make the model more stable when it has already reached the required feature. For XL models, we use a dropout of 0.1 on the middle third of the \texttt{SiT} blocks (both for attention and the feedforward layers). Implementation details are given in \cref{tab:impl_details}.

\paragraph{Results without guidance}
We first compare \df{\texttt{DF}} against the baseline \texttt{SiT} models in the unguided setting (\cref{tab:model_size_abla}). All three sizes improve substantially: \texttt{B}, \texttt{L} and \texttt{XL} all fall below 5 FID within 400K training steps (or 80 epochs), with XL reaching 2.61.

\begin{table}[!h]
\centering
\small
\caption{Model-size ablation with \texttt{DINO} as the frozen representation.}
\label{tab:model_size_abla}
\resizebox{\textwidth}{!}{%
\begin{NiceTabular}{@{}l ccc S[table-format=2.2,detect-weight,mode=text]@{}}
\toprule
\multicolumn{5}{c}{\textit{Latent space}} \\
\toprule
\textbf{Model} & \textbf{\#Params} & \textbf{Steps} & \textbf{Batch Size} & {\textbf{FID}\,$\downarrow$} \\
\midrule
\texttt{SiT-B/2}  & 130M & 400K & 256 & 33.00 \\
\rowcolor{dfcolor!7}\stripe{dfcolor}
\quad $\rightarrow$ \textbf{\texttt{DF}} & 138M & 400K & 256 & {\bfseries 3.66} \\
\greyrule
\texttt{SiT-L/2}  & 458M & 400K & 256 & 18.80 \\
\rowcolor{dfcolor!7}\stripe{dfcolor}
\quad $\rightarrow$ \textbf{\texttt{DF}} & 466M & 400K & 256 & {\bfseries 2.25} \\
\greyrule
\texttt{SiT-XL/2} & 675M & 400K & 256 & 17.20 \\
\rowcolor{dfcolor!7}\stripe{dfcolor}
\quad $\rightarrow$ \textbf{\texttt{DF}} & 684M & 400K & 256 & {\bfseries 1.93}  \\
\bottomrule
\end{NiceTabular}%
\hspace{2em}%
\begin{NiceTabular}{@{}l ccc S[table-format=2.2,detect-weight,mode=text]@{}}
\toprule
\multicolumn{5}{c}{\textit{Pixel space}} \\
\toprule
\textbf{Model} & \textbf{\#Params} & \textbf{Steps} & \textbf{Batch Size} & {\textbf{FID}\,$\downarrow$} \\
\midrule
\texttt{JiT-B/16} & 131M & 100K & 1024 & 48.61 \\
\rowcolor{dfcolor!7}\stripe{dfcolor}
\quad $\rightarrow$ \textbf{\texttt{DF}} & 139M & 100K & 1024 & \bfseries 13.37 \\
\greyrule
\texttt{JiT-L/16} & 459M & 100K & 1024 & 26.64 \\
\rowcolor{dfcolor!7}\stripe{dfcolor}
\quad $\rightarrow$ \textbf{\texttt{DF}} & 468M & 100K & 1024 & \bfseries 6.80 \\
\greyrule
\texttt{JiT-XL/16} & 676M & 100K & 1024 & 23.03 \\
\rowcolor{dfcolor!7}\stripe{dfcolor}
\quad $\rightarrow$ \textbf{\texttt{DF}} & 686M & 100K & 1024 & \bfseries 5.45 \\
\bottomrule
\end{NiceTabular}}
\end{table}

When compared with state-of-the-art works at a similar parameter count (\cref{fig:convergence}), \df{\texttt{DF}} reaches 1.93 at 80 epochs, better than methods trained substantially longer: \texttt{REPA} at 800 epochs (5.78), \texttt{CoReDi} at 400 epochs (3.30), \texttt{REG} at 200 epochs (2.70) and \texttt{REGLUE} at 200 epochs (2.50). When further trained for 20 more epochs \df{\texttt{DF} shows an $8\times$ reduction in training budget  (see \cref{tab:latent_end_abla}) over \texttt{REG} ($1.84$ after 100 epochs against $1.80$ after 800 epochs) and \texttt{RAE-DiT-XL} ($1.87$ after 800 epochs), the previous best methods at a comparable parameter count}. We must note that the excellent work, \texttt{TREAD}~\citep{krause2025tread}, remains competitive without using external pretrained model. Since the two approaches are orthogonal, combining \df{\texttt{DF}} with \texttt{TREAD} should yield further gains.

\begin{table}[!t]
    \centering
    \small
    \setlength{\tabcolsep}{4pt}
    \caption{State-of-the-art comparison of latent space models on ImageNet $256^2$. \textbf{Bold} is best, \underline{underline} is second best, {\color{gray}grayed methods} use more than 200 epochs and are not considered for the comparison.}
    \label{tab:latent_end_abla}
\resizebox{\textwidth}{!}{%
    \begin{NiceTabular}{@{}ll cc cccc cccc@{}}
    \toprule
     & & & & \Block{1-4}{\textbf{w/o CFG}} & & & & \Block{1-4}{\textbf{w/ CFG}} & & & \\
    \cmidrule(lr){5-8} \cmidrule(lr){9-12}
    \textbf{Model} & \textbf{DINO use} & \textbf{\#Params} & \textbf{Epochs} & FID\,$\downarrow$ & IS\,$\uparrow$ & P\,$\uparrow$ & R\,$\uparrow$
         & FID\,$\downarrow$ & IS\,$\uparrow$ & P\,$\uparrow$ & R\,$\uparrow$ \\
    \midrule
    \RowStyle{\color{gray}}\texttt{DiT}                             & None       & 675M & 1400 & 9.62 & 121.5 & 0.67 & 0.67 & 2.27 & 278.2 & 0.83 & 0.57 \\
    \RowStyle{\color{gray}}\texttt{SiT}                             & None       & 675M & 1400 & 8.61 & 131.7 & 0.68 & 0.67 & 2.06 & 270.3 & 0.82 & 0.59 \\
    \RowStyle{\color{gray}}\texttt{REPA}                            & Alignment  & 675M &  800 & 5.78 & 158.3 & 0.70 & 0.68 & 1.42 & 305.7 & 0.80 & 0.65 \\
    \RowStyle{\color{gray}}\texttt{RAE-DiT\textsuperscript{DH}-XL}  & Rep.\ space & 839M &  800 & 1.51 & 242.9 & 0.79 & 0.63 & 1.13 & 262.6 & 0.78 & 0.67 \\
    \RowStyle{\color{gray}}\texttt{RAE-DiT-XL}                      & Rep.\ space & 676M &  800 & 1.87 & 209.7 & 0.80 & 0.63 & 1.41 & 309.4 & 0.80 & 0.63 \\
    \RowStyle{\color{gray}}\texttt{ReDi}                            & Joint gen. & 675M &  800 & --   & --    & --   & --   & 1.61 & 295.1 & 0.78 & 0.64 \\
    \RowStyle{\color{gray}}\texttt{REG}                             & Joint gen. & 677M &  800 & 1.80 & --    & --   & --   & 1.36 & 299.4 & 0.77 & 0.66 \\
    \midrule
    \RowStyle{\color{gray}}\texttt{DiT + TREAD}                     & None       & 675M &  680 & 3.93 & 211.4 & 0.76 & 0.60 & --   & --    & --   & --   \\
    \RowStyle{\color{gray}}\texttt{HASTE}                           & Alignment  & 675M &  500 & --   & --    & --   & --   & 1.42 & 299.5 & 0.80 & 0.65 \\
    \RowStyle{\color{gray}}\texttt{DDT}                             & Alignment  & 675M &  400 & 6.27 & 154.7 & 0.68 & 0.69 & 1.26 & 310.6 & 0.79 & 0.65 \\
    \RowStyle{\color{gray}}\texttt{CoReDi}                          & Joint gen. & 675M &  400 & 3.30 & 176.8 & 0.74 & 0.66 & 1.58 & 297.2 & 0.63 & 0.78 \\
    \midrule
    \texttt{REED}                         & Alignment  & 675M &  200 & 4.70 & --    & --   & --   & \underline{1.80} & \underline{267.5} & \underline{0.81} & 0.61 \\
    \texttt{REG}                          & Joint gen. & 677M &  200 & 2.70 & --    & --   & --   & --   & --    & --   & --   \\
    \texttt{SFD+LDiT/1}                   & Joint gen. & 675M &  200 & 2.54 & \underline{191.7} & 0.75 & \underline{0.67} & \textbf{1.06} & 267.0 & 0.78 & \underline{0.67} \\
    \texttt{REGLUE}                       & Joint gen. & 677M &  200 & \underline{2.50} & 188.6 & \underline{0.76} & 0.65 & --   & --    & --   & --   \\
    \midrule
    \texttt{iREPA}                        & Alignment  & 675M &   80 & 7.13 & 132.9 & 0.71 & 0.66 & 1.89 & \textbf{272.4} & 0.80 & 0.60 \\
    \midrule
    \rowcolor{dfcolor!7}\stripe{dfcolor} \ \ \textbf{\texttt{DF}} & Forcing & 684M & 100 & \textbf{1.84} & \textbf{259.2} & \textbf{0.85}  & \textbf{0.69} & 1.82 & 264.2 & \textbf{0.85} & \textbf{0.69} \\
    \bottomrule
    \end{NiceTabular}}
\end{table}

\paragraph{Results with guidance}

Like \texttt{RAE}~\citep{zheng2026diffusion}, we found guidance non-trivial to implement for our latent space model. We discuss further the reasoning behind this failure in Appendix~\ref{app:cfg_fail}. 
As such, following \texttt{RAE} we also use Autoguidance~\citep{karras2024guiding} and use an \texttt{L} model's checkpoint at 150K steps to act as the weaker model. This leads to a guided FID score of 1.82 (see \cref{tab:latent_end_abla}). It reflects an improvement; however its not as drastic as prior works, which are trained for at least 2$\times$ more than our setup. We agree that the guided scores are still a bit underwhelming and more work must be done to improve on it. Interesting ideas like \texttt{Internal Guidance}~\citep{zhou2026guiding} with intermediate outputs can be an interesting avenue to explore to improve upon the guidance, and we leave this as future work to further improve upon our core idea.

\paragraph{High resolution}
We next evaluate at $512\times 512$ resolution in the unguided setting (\cref{tab:high_res}). \df{\texttt{DF}} reaches an FID of $3.11$ at 80 epochs, against $11.05$ for \texttt{REPA} and $20.45$ for the \texttt{SiT-XL/2} baseline. The Inception Score improves from $116.2$ to $274.5$. Precision is unchanged ($0.65$), but recall drops from $0.73$ to $0.63$, indicating the predicted representation condition concentrates the sample distribution: it acts on the trajectory much as guidance would, consistent with the large unguided FID gap.

\begin{table}[t]
    \centering
    \small
    \setlength{\tabcolsep}{4pt}
    \caption{State-of-the-art comparison of pixel-space models on ImageNet $256^2$. \textbf{Bold} is best, \underline{underline} is second best, {\color{gray}grayed methods} use more than 200 epochs and are not considered for the comparison.}
    \label{tab:pixel_abla}
\resizebox{\textwidth}{!}{%
    \begin{NiceTabular}{@{}ll cc rrcc rrcc@{}}
    \toprule
     & & & & \Block{1-4}{\textbf{w/o CFG}} & & & & \Block{1-4}{\textbf{w/ CFG}} & & & \\
    \cmidrule(lr){5-8} \cmidrule(lr){9-12}
    \textbf{Model} & \textbf{DINO use} & \textbf{\#Params} & \textbf{Epochs} & FID\,$\downarrow$ & IS\,$\uparrow$ & P\,$\uparrow$ & R\,$\uparrow$
         & FID\,$\downarrow$ & IS\,$\uparrow$ & P\,$\uparrow$ & R\,$\uparrow$ \\
    \midrule
    \RowStyle{\color{gray}}\texttt{JiT-G}                 & None       & 2B   & 600 & --    & --    & --   & --   & 1.82 & 292.6 & --   & --   \\
    \RowStyle{\color{gray}}\texttt{JiT-XL + $fv$-loss}    & None       & 676M & 600 & 11.5  & 118.5 & --   & --   & 2.13 & 290.3 & --   & --   \\
    \RowStyle{\color{gray}}\texttt{DeCo-XL}               & Alignment  & 682M & 600 & --    & --    & --   & --   & 1.69 & 304.0 & 0.79 & 0.63 \\
    \RowStyle{\color{gray}}\texttt{PixelREPA}             & Alignment  & 953M & 600 & --    & --    & --   & --   & 1.81 & 317.2 & --   & --   \\
    \RowStyle{\color{gray}}\texttt{PixelDiT2-H}           & Grounding  & 1B   & 480 & --    & --    & --   & --   & 1.48 & 299.1 & --   & --   \\
    \midrule
    \RowStyle{\color{gray}}\texttt{PixelDiT-XL}           & Alignment  & 797M & 320 & --    & --    & --   & --   & 1.61 & 292.7 & 0.78 & 0.64 \\
    \RowStyle{\color{gray}}\texttt{V-Co}                  & Joint gen. & 1.9B & 300 & --    & --    & --   & --   & 1.71 & 263.3 & --   & --   \\
    \midrule
    \texttt{JiT-L}                        & None       & 459M & 200 & 17.19 & 85.4  & 0.70 & \textbf{0.73} & 3.06 & 271.5 & \textbf{0.85} & 0.66 \\
    \texttt{JiT-XL}                       & None       & 676M & 200 & 12.71 & 101.7 & \underline{0.74} & \textbf{0.73} & 2.89 & 272.1 & \textbf{0.85} & 0.66 \\
    \texttt{LatentForcing-L}              & Joint gen. & 465M & 200 & 7.20  & --    & --   & --   & 2.48 & --    & --   & --   \\
    \texttt{PixelDiT2-L}                  & Grounding  & 503M & 200 & --    & --    & --   & --   & 2.35 & 270.1 & --   & --   \\
    \texttt{PixelGen-XL}                  & Percept.+Align. & 676M & 160 & --    & --    & --   & --   & \textbf{1.83} & 293.6 & 0.79 & 0.63 \\
    \midrule
    \rowcolor{dfcolor!7}\stripe{dfcolor} \ \
    \textbf{\texttt{DF-L (ours)}}         & Forcing    & 468M & {200} & \underline{5.53} & \underline{163.6} & \textbf{0.79} & \underline{0.72} & 2.26 & \underline{309.9} & \underline{0.84} & \underline{0.70} \\
    \rowcolor{dfcolor!7}\stripe{dfcolor} \ \
    \textbf{\texttt{DF-XL (ours)}}        & Forcing    & 686M & {200} & \textbf{4.56} & \textbf{179.6} & \textbf{0.79} & \textbf{0.73} & \underline{2.13} & \textbf{310.0} & 0.83 & \textbf{0.71} \\
    \bottomrule
    \end{NiceTabular}}
\end{table}

\begin{figure}[t]
    \centering
    \begin{minipage}[t]{0.58\textwidth}
        \vspace{0pt}
        \centering
        \input{images/dino_inversion}
        \caption{Reconstructions from inverted \texttt{DINOv2-B} features.
                 \textbf{Top} row: originals; \textbf{bottom} row: reconstructions.}
        \label{fig:dino_inversion}
    \end{minipage}\hfill
    \begin{minipage}[t]{0.4\textwidth}
        \vspace{0pt}
        \centering
        \small
        \setlength{\tabcolsep}{2pt}

        \captionof{table}{Comparison on ImageNet $512^2$.
        without classifier-free guidance.}
        \label{tab:high_res}
        \vspace{-3mm}
        \begin{NiceTabular}{@{}l rrcc@{}}
            \toprule
            \textbf{Model} & FID\,$\downarrow$ & IS\,$\uparrow$
                           & P\,$\uparrow$ & R\,$\uparrow$ \\
            \midrule
            \texttt{SiT-XL/2} & 20.45 & 75.1  & \textbf{0.66} & 0.68 \\
            \texttt{REPA}     & 11.05 & 116.2 & 0.65 & \textbf{0.73} \\
            \rowcolor{dfcolor!7}\stripe{dfcolor} \ 
            \textbf{\texttt{DF (ours)}} & \textbf{3.11}
                & \textbf{274.5} & 0.65 & 0.63 \\
            \bottomrule
        \end{NiceTabular}

        \centering
\small
\setlength{\tabcolsep}{4pt}
\captionof{table}{DINO inversion quality.}
\label{tab:psnr_ssim}
\begin{NiceTabular}{@{}l cc@{}}
  \toprule
  \textbf{Model} & PSNR\,$\uparrow$ & SSIM\,$\uparrow$ \\
  \midrule
  \texttt{DF-SiT-XL} & 16.86 & 0.48 \\
  \texttt{DF-JiT-XL} & 14.33 & 0.44 \\
  \bottomrule
\end{NiceTabular}
    \end{minipage}
\end{figure}

\subsection{Pixel Space Setting}
\label{sec:pixelspace}

\paragraph{Setup} For pixel space, we closely follow the implementations of \texttt{JiT}~\citep{li2026back} and we use \texttt{DINOv2-B} as the pretrained representation. We observe that $v$-loss in pixel-space is $~13\times$ lower than in latent space so we lower $\lambda = 0.03$ in \cref{eq:loss}. We use Adam optimizer without any weight decay, with a constant learning rate of $2e-4$ after a warmup for 5 epochs. For pixel space experiments, we follow Heun Sampler~\citep{heun1900neue} with 50 sampling steps. Similar to the latent space setting, we use the null condition 10\% of the training, optimize on the first forward pass 10\% of the training, use the original \texttt{DINOv2-B} features for 5\% of the training, and we use a dropout rate of $0.1$ for our bigger XL models. Additional implementation details are further provided in \cref{tab:impl_details}. 

\paragraph{Results without guidance}
The improvements observed with \texttt{SiT} carry over to \texttt{JiT} (see \cref{tab:model_size_abla}): all model sizes improve with \texttt{XL} reaching an unguided FID of $5.45$. We compare \df{\texttt{DF}} againt \texttt{Latent Forcing}~\citep{baade2026latent} in \cref{tab:pixel_abla}. At 200 epochs, \df{\texttt{DF}} reaches $5.53$ with an L-sized model, against $7.20$. We attribute this gap to how the pretrained representation signal is produced: \texttt{Latent Forcing} commits to a single schedule, fixed at training time and shared across the dataset, that dictates when pretrained representation information is revealed relative to pixels. \df{\texttt{DF}} instead derives the conditioning from the sample being generated, so the pretrained representation signal adapts to each trajectory rather than following a common ordering.

\paragraph{Results with guidance}

Prior co-denoising works in pixel space, like \texttt{Latent Forcing}~\citep{baade2026latent}, require a split scheme, applying AutoGuidance~\citep{karras2024guiding} on pixel timesteps and CFG interval~\citep{kynkaanniemi2024applying} on representation timesteps, because the two modalities advance on separate schedules. Having a single trajectory, we need no such split: the default interval guidance setting works well without method-specific tuning (\cref{tab:pixel_abla}). At \texttt{L} size we reach 2.26, surpassing \texttt{Latent Forcing} (2.48) as well as \texttt{PixelDiT-2} (2.35). At \texttt{XL} size we reach 2.13, against 2.89 for \texttt{JiT-XL} under the same setting. We remain behind \texttt{PixelGen-XL}, which reaches 1.83 at 160 epochs in a similar parameter range; we attribute this to its use of frozen \texttt{VGG}~\citep{simonyan2014very} supervision alongside \texttt{DINO}, and leave the combination of the two signals to future work.

\subsection{DINO inversion}
\label{sec:dino_inversion}

Our method assumes the branch produces something semantically meaningful, not only an arbitrary code that happens to help the velocity prediction. We visualise the PCA of the predicted features across the trajectory in \cref{fig:codenoising} and \cref{fig:qual_dino_progression}): their structures are spatially coherent and separates object from background before the image itself is resolved, and it sharpens as denoising progresses. 

Furthermore, as the representation-conditioning branch processes representation features, we can bypass the prediction at inference and feed real \texttt{DINOv2} features instead. This turns the model into a \texttt{DINOv2} inverter: given the features of an image and a null class, it reconstructs the image itself. \cref{fig:dino_inversion} shows reconstructions from inverted \texttt{DINOv2}. Table \ref{tab:psnr_ssim} reports SSIM and PSNR for \texttt{SiT} (16.86 dB / 0.48 SSIM) and \texttt{JiT} (14.33 dB / 0.44 SSIM), showing that the interaction of the \texttt{DINOv2} condition provides very good local information for reconstruction.

\subsection{Generation without class condition}
\label{sec:dino_cond_only}

The predicted \texttt{DINOv2} from the representation-conditioning branch can be seen as a standalone condition. To test this, we train with the class token set to null for the entire run, so the model has no conditioning signal other than the one it predicts for itself. We compare this setup with unconditional \texttt{SiT} and \texttt{JiT} and with \texttt{Latent Forcing}~\citep{baade2026latent} in \cref{tab:unconditional_gen}. In latent space, \df{\texttt{DF}} reaches an FID of $9.14$ against $49.51$ for \texttt{SiT-L/2}. In pixel space it reaches $15.19$, against $53.97$ for \texttt{JiT-L/16} and $20.44$ for \texttt{Latent Forcing}. This setting isolates the main difference with \texttt{Latent Forcing}: when all of the conditioning comes from the pretrained representation branch, predicting it outperforms denoising it on a fixed schedule.


\begin{table}[t]
\centering

\begin{minipage}[t]{0.49\textwidth}
    \vspace{0pt}
    \centering
    \small
    \setlength{\tabcolsep}{3pt}
    \captionof{table}{Unconditional generation on ImageNet $256^2$.
    (no CFG, 80 epochs). }
    \label{tab:unconditional_gen}
    \resizebox{\textwidth}{!}{%
    \begin{NiceTabular}{@{}l ccr@{}}
        \toprule
        \textbf{Model} & \textbf{\#Params} & \textbf{Steps}
        & \textbf{FID}\,$\downarrow$ \\
        \midrule
        \texttt{SiT-L/2} & 458M & 400K &  49.51 \\
        \rowcolor{dfcolor!7}\stripe{dfcolor}\quad $\rightarrow$ \textbf{\texttt{DF}}
            & 466M & 400K &  \textbf{8.19} \\
        \midrule
        \texttt{JiT-L/16} & 459M & 100K  & 53.97 \\
        \quad $\rightarrow$ \texttt{LatentForcing}
            & 465M & 100K & 20.44 \\
        \rowcolor{dfcolor!7}\stripe{dfcolor}\quad $\rightarrow$ \textbf{\texttt{DF}}
            & 468M & 100K  & \textbf{15.19} \\
        \bottomrule
    \end{NiceTabular}}
\end{minipage}\hfill%
\begin{minipage}[t]{0.49\textwidth}
    \vspace{0pt}
    \centering
    \small
    \setlength{\tabcolsep}{3pt}
    
    \captionof{table}{Effect of the DF supervision target on SiT-L/2
    at 80 epochs. }
    \label{tab:latent_self_cond_abla}
    
      \resizebox{\textwidth}{!}{%
    \begin{NiceTabular}{@{}l cr@{}}
        \toprule
        \textbf{Model} & \textbf{\#Params} 
        & \textbf{FID}\,$\downarrow$ \\
        \midrule
        \texttt{SiT-L/2} & 458M &  18.80 \\
        \quad $\rightarrow$ \texttt{DF}-$\emptyset$
            & 466M & 12.96 \\
        \quad $\rightarrow$ \texttt{DF}-image
            & 466M &  11.93 \\
        \quad $\rightarrow$ \texttt{DF}-\texttt{MAE}
            & 466M & 7.97 \\
        \rowcolor{dfcolor!7}\stripe{dfcolor}\quad $\rightarrow$
        \textbf{\texttt{DF}-\texttt{DINO} supervision}
            & 466M & \textbf{2.25} \\
        \bottomrule
    \end{NiceTabular}}
    
\end{minipage}

\end{table}

\subsection{Ablation}
\label{sec:ablation}

We ablate design choices of \mname; more ablations in Appendix~\ref{app:sec:ablation}. Unless stated otherwise, every ablation is run on \texttt{SiT-L} and reported in the unguided setting, so that all numbers are directly comparable.

\paragraph{Does the pretrained representation target matter?} The improvement from \df{\texttt{DF}} could come from the conditioning mechanism, or from the content of the frozen encoder. To separate the two, we train four variants with identical architectures that differ only in what the representation branch is asked to predict (\cref{tab:latent_self_cond_abla}). \texttt{DF}-$\emptyset$ leaves the predicted features unsupervised. \texttt{DF}-image supervises the branch to predict the image itself, a well-defined but low-level target. \texttt{DF}-\texttt{MAE} uses Masked Autoencoder (\texttt{MAE-B})~\citep{mae} features, a strong reconstruction-based self-supervised target. \texttt{DF}-\texttt{DINO} supervision uses \texttt{DINOv2} features, as in our full method. The conditioning mechanism helps on its own: without any supervision, FID drops from $18.80$ to $12.96$. Supervising with the raw image adds little ($11.93$). So a reconstruction target is not what drives the gain. \texttt{MAE} features show much broader improvement (more than a 10-point improvement over baseline). However, with \texttt{DINOv2} features, FID reaches $2.25$, $9 \times$ below the baseline. The conditioning mechanism is necessary but not sufficient; the pretrained representation content of the target is what produces most of the improvement.

\subsection{\texorpdfstring{Training cost}{Training cost}}
\label{sec:training_cost}
 
\begin{table}[t]
\centering

\begin{minipage}[t]{0.48\textwidth}
    \vspace{0pt}
    \centering
    \small
    \setlength{\tabcolsep}{3pt}
    \captionsetup{position=top}
    \captionof{table}{Training compute of \texttt{SiT-XL} on ImageNet $256^2$ (no CFG). PFLOPs include \texttt{DINOv2} extraction and both \texttt{DF} passes.}
    \label{tab:train_flops}
    \begin{NiceTabular}{@{}w{l}{1.45cm} w{c}{0.85cm} w{r}{1.75cm} w{r}{1.15cm}@{}}
        \toprule
        \textbf{Model} & \textbf{Steps} & \textbf{PFLOPs}\,$\downarrow$
        & \textbf{FID}\,$\downarrow$ \\
        \midrule
        \texttt{REG}    & 1M & 197,493 & 2.70 \\
        \texttt{REGLUE} & 1M & 198,812 & 2.50 \\
        \midrule
        \rowcolor{dfcolor!7}\stripe{dfcolor} \ \ \textbf{\texttt{DF}}
            & 500K & 126,678 & \textbf{1.84} \\
        \bottomrule
    \end{NiceTabular}
\end{minipage}\hfill%
\begin{minipage}[t]{0.48\textwidth}
    \vspace{0pt}
    \centering
    \small
    \setlength{\tabcolsep}{3pt}
    \captionsetup{position=top}
    \captionof{table}{Wall-clock training time of \texttt{SiT-XL} on ImageNet $256^2$ (no CFG), in H100 GPU-hours. Same hardware for all methods, batch size 256.}
    \label{tab:train_time}
    \begin{NiceTabular}{@{}w{l}{1.45cm} w{c}{0.85cm} w{r}{1.75cm} w{r}{1.15cm}@{}}
        \toprule
        \textbf{Model} & \textbf{Steps} & \textbf{GPU-hours}\,$\downarrow$
        & \textbf{FID}\,$\downarrow$ \\
        \midrule
        \texttt{REG} & 1M   & 228.9 & 2.70 \\
        \texttt{REGLUE}  & 1M & 233.6 & 2.50 \\
        \midrule
        \rowcolor{dfcolor!7}\stripe{dfcolor} \ \ \textbf{\texttt{DF}}
            & 500K & \textbf{146.1} & \textbf{1.84} \\
        \bottomrule
    \end{NiceTabular}
\end{minipage}

\end{table}
 
\mname{} is highly data efficient and achieves better unguided FID scores in 8$\times$ less number of epochs in latent space setting (see \cref{tab:latent_end_abla}). However, during training, as noted in \cref{sec:method}, \mname{} involves an additional gradient free look ahead pass (see Pass-1 in \cref{fig:arch}). We quantify the effect of this extra compute w.r.t. FLOPs and wall-clock training time.

 
\paragraph{Training FLOPs} We adopt the accounting reported in \texttt{REGLUE}~\citep{petsangourakis2025reglue} (Table 13 in the \texttt{REGLUE} paper) and calculate training FLOPs by including the feature extraction of pretrained encoder. We report the values in \cref{tab:train_flops}. Despite having an extra gradient-free forward pass (which is typically cheap compared to backward pass), \mname{} reaches an unguided FID of $1.84$ in just $127$K PFLOPs, outperforming both \texttt{REG}~\citep{wu2026representation} ($2.70$ FID with $197$K PFLOPs) and \texttt{REGLUE} ($2.50$ FID with $199$K PFLOPs); a substantial $36\%$ less training compute usage but still achieving far superior performance.
 
\paragraph{Wall-clock time} Apart from compute FLOPs we also compare training time using an NVIDIA H100 GPU for \mname{} and compare it with prior works. We report the results in \cref{tab:train_time} for latent space setting. \mname{} reaches an unguided FID of $1.84$ in just $146.1$ GPU-hours, which is $36.2\%$ less time than \texttt{REG} (FID $2.70$ in $228.9$ hours) and $1.6\times$ faster than \texttt{REGLUE} (FID $2.50$ in $233.6$ hours) while still having better FID scores.
\section{Conclusion}
\label{sec:discussion}

We introduced \texttt{DINO-Forcing}, a simple alternative to co-denoising pretrained representations in flow-matching models. Instead of assigning the representation its own noisy trajectory, we predict it from the current image state and immediately reuse it as conditioning. This removes the second ODE and representation-specific denoising schedule, while preserving the benefits of representation guidance.
Despite its simplicity, this change substantially improves both convergence and generation quality, surpassing methods trained for substantially longer both in latent and pixel space. More broadly, our results suggest that auxiliary representations need not always be generated alongside the primary modality: when they can be inferred from the evolving sample, prediction may be enough. 

\subsection*{Acknowledgement}
First and foremost, the authors would like to thank \textbf{Nicolas Dufour}, for all the productive discussions and arguments in the lab, in David's office, in pubs, and in Slack! The authors would also like to thank \textbf{Felix Krause} from LMU, for discussing and sharing checkpoints of his amazing work, \texttt{TREAD}~\citep{krause2025tread}. It played a tremendous role in helping the authors find a good and easy solution to the problem at hand, even though the final paper did diverge quite far from what \texttt{TREAD} proposed. The authors would like to thank \textbf{Zeynep Sonat Baltacı}, \textbf{Gatien Chenu}, \textbf{Fei Meng} and \textbf{Julie Mordacq} for taking a look at the first draft of the paper (with some even doing this during their amazing trip in the mountains), proofreading like \textit{Sherlock}, and also providing cool feedback! The authors would also like to thank \textbf{Yohann Perron}, \textbf{Alexandros Benetatos}, \textbf{Loic Landrieu}, \textbf{Tristan Quétin}, \textbf{Tom Ravaud}, \textbf{Louis Geist} and \textbf{Lucas Ventura} for many insightful discussions at many different times while working on this paper! Finally, the authors would like to thank the city of Paris and its beautiful summer this year (skipping the heatwaves) for being highly motivating during the preparation of this paper. 

\paragraph{Funding and Compute} This work was supported by a Hi!Paris grant, two Hi!Paris chairs for A.A.Efros and
V.Kalogeiton, ANR/France 2030 program (ANR-23-IACL-0005) and ANR project sharp ANR-23-PEIA-0008 in the context of the PEPR IA. V.Kalogeiton is co-funded by the European Union (ERC, FLASH, 101303557). 
Regarding compute, the paper was granted access to the HPC resources of IDRIS under the allocations 2025-A0181016194, 2025-AD011015436, 2026-A0201017545 and 2026-AD011015594R2 made by GENCI. 

\subsection*{AI use statement}

In this work, we used generative AI tools to design or provide feedback on research methodology or experiments. We have not used generative AI tools to help develop theoretical models or conceptual frameworks, and the rest of the required disclosure tasks (generate synthetic data sets, formulate mathematical claims, provide critical ingredients for proving mathematical claims, assist in the writing of proofs, propose or refine hypotheses, implement methods, assist with translation, clean and reformat dataset, support qualitative and thematic data analysis, interpret results) are not applicable to this work. 

Additionally, we used generative AI tools to create or modify scientific figures or images, creation of artifacts, draft parts of a research paper, brainstorming, sourcing/searching for information, edit a research paper to improve readability, identify relevant literature, and format references.

We have reviewed all AI-assisted work. We checked LLM-generated code, LLM-generated suggestions have been tested and backed-up by experimental results. We take responsibility for the final content of this work, including text, claims or artifacts produced with the aid of generative AI.

\bibliography{iclr2027_conference}
\bibliographystyle{iclr2027_conference}

\newpage
\appendix
\section*{Appendix}

\section{Implementation Details}
\label{subsec:appendix_impl_details}

\begin{table}[!h]
    \centering
    \small
    \setlength{\tabcolsep}{6pt}
    \newcommand{\implgroup}[1]{\multicolumn{3}{@{}l}{\textit{#1}}\\}
    \caption{Training and sampling configuration of \df{\texttt{DF}} in latent and pixel space on ImageNet.}
    \label{tab:impl_details}
    \begin{NiceTabular}{@{}l cc@{}}
    \toprule
    & \textbf{Latent space} & \textbf{Pixel space} \\
    \midrule
    \implgroup{Backbone}
    Base implementation            & \texttt{SiT}~\citep{ma2024sit}          & \texttt{JiT}~\citep{li2026back} \\
    Input space                    & \texttt{sd-vae-ft-ema} latents          & RGB pixels \\
    Resolution                     & $256\times256$, $512\times512$          & $256\times256$ \\
    Patch size                     & 2                                       & 16 \\
    Dropout (XL only)              & 0.1, middle third of blocks             & 0.1, middle third of blocks \\
    \midrule
    \implgroup{Representation Branch}
    Target encoder $\phi$          & \texttt{DINOv2-B}                       & \texttt{DINOv2-B} \\
    Representation dim.\ $D'$      & 768                                     & 768 \\
    Fusion                         & channel concat.\ + linear               & channel concat.\ + linear \\
    Prediction block index              & last                                    & last \\
    Prediction block type              & \texttt{REPA MLP}                                    & \texttt{REPA MLP} \\
    $\mathcal{L}_{\text{FR}}$      & cosine similarity                       & cosine similarity \\
    Loss weight $\lambda$          & 0.5                                     & 0.03 \\
    \midrule
    \implgroup{Optimization}
    Optimizer                      & Adam                                    & Adam \\
    Learning rate                  & $1\times10^{-4}$ (constant)             & $2\times10^{-4}$ (constant) \\
    Warmup                         & --                                      & 5 epochs \\
    Weight decay                   & 0                                       & 0 \\
    Batch size                     & 256                                     & 1024 \\
    EMA decay                      & 0.9999                                  & 0.9999 \\
    Training epochs                & 80 (\texttt{B}, \texttt{L}) / 100 (\texttt{XL})                                & 80 (\texttt{B}) / 200 (\texttt{L}, \texttt{XL}) \\
    \midrule
    \implgroup{Conditioning Dropout}
    Null class (CFG)               & 10\%                                    & 10\% \\
    Pass~1 only ($p$)              & 10\%                                    & 10\% \\
    Ground-truth $\phi(x)$ as condition & 5\%                                & 5\% \\
    \midrule
    \implgroup{Sampling}
    Solver                         & Euler (ODE)                             & Heun (ODE) \\
    Steps                          & 50                                      & 50 \\
    Guidance                       & AutoGuidance (\texttt{L} @ 150K)        & CFG + interval (0.1,1) \\
    CFG sweep range & (1.0, 4.0) & (1.0, 4.0) \\
    CFG sweep range interval & 0.1 & 0.1 \\
    \bottomrule
    \end{NiceTabular}
\end{table}

\section{Explanation behind our fusing Image and Frozen Representations mechanism}
\label{app:sub:fuse}
One particularly interesting question that arises from \cref{fig:arch} is how we \textit{merge} the image and the frozen representations. Given we use \texttt{DiT} based transformer architecture, we use same number of tokens as in the image. As such, any token-wise merge operation suffices. However, for our experiments, we fuse our image and representation tokens by concatenating them via the channel dimension. Given an image (or its feature), $x$, we extract patch tokens of shape \texttt{(B,T,D)} via standard patchification. We set our starting representation token dimension to \texttt{(B,T,D')} matching the dimension of frozen pretrained encoder (e.g., \texttt{DINOv2}~\citep{oquab2024dinov}). The two token sequences are thereafter spatially aligned and concatenated along the channel dimension, yielding a fused representation of shape \texttt{(B,T,D+D')}, which is then projected to the transformer's hidden dimension via a lightweight linear layer. This keeps the token count identical to a standard image-only model and requires no spatial resampling, making it directly compatible with existing flow matching architectures. Our fusion strategy is similar in spirit to concurrent work~\citep{kouzelis2026boosting}, but simpler by virtue of requiring no upsampling. There are many other ways we can fuse the tokens, and we urge the readers to follow the excellent \texttt{V-Co}~\citep{lin2026v} paper, which does a deep dive into different fusion techniques. Here, we stick to the channel concatenation strategy just for proper comparison and leave the exploration of other types of fusion strategies for future work.

\section{Intuition behind the failure in classifier-free guidance in latent space setting}
\label{app:cfg_fail}

Similar to prior works, like \texttt{RAE}~\citep{zheng2026diffusion}, we also found it difficult to adapt classifier-free guidance for our latent space model. We hypothesize the main reason for this was the \texttt{DINOv2} self-condition, which is too strong. To test our hypothesis, we set up a small experiment. We train a small linear probing network using $100$K true \texttt{DINOv2} features. It achieved a top-1 accuracy of 89.1\% on a small held-out set of $2$K features. We then used our latent space \mname{} model to generate $2$K \texttt{DINOv2} features (2 features per class of ImageNet) with and without classifier-free guidance. Then we passed these features through the linear probing model to recover the class labels. We found that the recovered class labels had a top-1 accuracy of 89.1\% for the unguided features when compared with the class labels used for generating those features; matching the original linear probing accuracy. However, the features generated using guidance surpassed that and achieved a top-1 accuracy of 96\%; clearly suggesting that the \texttt{DINOv2} self-condition features already hold most of the class information and as such classifier-free guidance doesn't bring much gain.

\section{Ablation}
\label{app:sec:ablation}

\begin{figure*}[t]
    \centering
    \resizebox{\textwidth}{!}{\begingroup
\renewcommand\familydefault{\sfdefault}

\definecolor{layer7} {RGB}{222,208,238}
\definecolor{layer8} {RGB}{200,182,226}
\definecolor{layer9} {RGB}{178,156,212}
\definecolor{layer12}{RGB}{156,130,196}
\definecolor{layer15}{RGB}{134,106,178}
\definecolor{layer18}{RGB}{112, 84,158}
\definecolor{drop0.5}{RGB}{200,182,226}
\definecolor{drop0.2}{RGB}{134,106,178}


\pgfplotsset{
  curve/.style={
    line width=2.4pt, line join=round, line cap=round,
    mark=*, mark size=2.3pt,
    mark options={solid, draw=white, line width=1.1pt},
  },
  fidpanel/.style={
    width=6.3cm, height=5.7cm,
    xlabel={Training steps},
    ylabel={FID $\downarrow$},
    label style={font=\normalsize},
    tick label style={font=\small},
    title style={font=\normalsize\bfseries, yshift=-2pt},
    xmin=70000, xmax=265000,
    xtick={75000,100000,150000,200000,250000},
    xticklabels={75K,100K,150K,200K,250K},
    ytick={2.5,5,10,20},
    yticklabels={2.5,5,10,20},
    minor tick num=0,
    log ticks with fixed point,
    grid=major,
    major grid style={draw=black!8, line width=0.4pt},
    axis line style={draw=black!50, line width=0.6pt},
    tick style={draw=black!50},
    axis lines=left,
    enlargelimits=false,
    unbounded coords=discard,
    every axis plot/.append style={curve},
    legend cell align=left,
    legend style={
      draw=none, fill=white, fill opacity=0.85, text opacity=1,
      font=\small, row sep=-1pt,
      /tikz/every even column/.append style={column sep=3pt},
    },
    legend image code/.code={\draw[mark repeat=2, mark phase=2, #1]
      plot coordinates {(0cm,0cm) (0.3cm,0cm) (0.6cm,0cm)};},
  },
}

\tikzset{dflbl/.style={font=\small\bfseries, text=dfcolor!70!black,
  inner sep=1.5pt, anchor=north, yshift=-14pt}}

\begin{tikzpicture}
\begin{loglogaxis}[fidpanel, title={Target layer}, ymin=2.2, ymax=30,
  legend columns=2, legend style={at={(0.98,0.98)}, anchor=north east}]
\addplot[layer7] coordinates {
  (75000,14.2145) (100000,10.0696) (125000,8.0503) (150000,6.9263) (175000,6.2278) (200000,5.6875) (225000,5.3192) (250000,5.0916)
  };
\addlegendentry{7}
\addplot[layer8] coordinates {
  (75000,12.7562) (100000,8.9627) (125000,7.3326) (150000,6.3612) (175000,5.7836) (200000,5.3508) (225000,5.0311) (250000,4.8077)
  };
\addlegendentry{8}
\addplot[layer9] coordinates {
  (75000,12.3141) (100000,8.5556) (125000,6.9112) (150000,5.9610) (175000,5.2805) (200000,4.8998) (225000,4.6179) (250000,4.4320)
  };
\addlegendentry{9}
\addplot[layer12] coordinates {
  (75000,11.8156) (100000,8.2343) (125000,6.6738) (150000,5.5966) (175000,5.0386) (200000,4.5962) (225000,4.3142) (250000,4.1823)
  };
\addlegendentry{12}
\addplot[layer15] coordinates {
  (75000,11.9555) (100000,8.2802) (125000,6.5851) (150000,5.5831) (175000,4.9755) (200000,4.5342) (225000,4.3459) (250000,4.1686)
  };
\addlegendentry{15}
\addplot[layer18] coordinates {
  (75000,9.9514) (100000,6.6673) (125000,5.2458) (150000,4.4428) (175000,3.9712) (200000,3.6264) (225000,3.3955) (250000,3.2575)
  };
\addlegendentry{18}
\addplot[dfcolor] coordinates {
  (75000,7.2341) (100000,4.9087) (125000,3.9260) (150000,3.3660) (175000,3.0378) (200000,2.8417) (225000,2.6869) (250000,2.6369)
  };
\addlegendentry{23}
\node[dflbl] at (axis cs:150000,4.1317) {DF};
\end{loglogaxis}
\end{tikzpicture}%
\hspace{0.3cm}%
\begin{tikzpicture}
\begin{loglogaxis}[fidpanel, title={Drop rate}, ymin=2.2, ymax=30,
  legend style={at={(0.98,0.98)}, anchor=north east}]
\addplot[drop0.5] coordinates {
  (100000,13.9779) (125000,10.9042) (150000,9.0831) (175000,7.7873) (200000,7.0661) (225000,6.3012) (250000,5.7400) 
  };
\addlegendentry{drop 0.5}
\addplot[drop0.2] coordinates {
  (75000,9.3694) (100000,6.1008) (125000,4.6013) (150000,3.8578) (175000,3.3652) (200000,3.0522) (225000,2.9242) (250000,2.7299) 
  };
\addlegendentry{drop 0.2}
\addplot[dfcolor] coordinates {
  (75000,7.2341) (100000,4.9087) (125000,3.9260) (150000,3.3660) (175000,3.0378) (200000,2.8417) (225000,2.6869) (250000,2.6369)
  };
\addlegendentry{drop 0.1}
\node[dflbl] at (axis cs:150000,4.1317) {DF};
\end{loglogaxis}
\end{tikzpicture}
\endgroup}
    \caption{Layer index and dropping probability ablation on SiT-L model.}
    \label{fig:layer_index_abla}
\end{figure*}
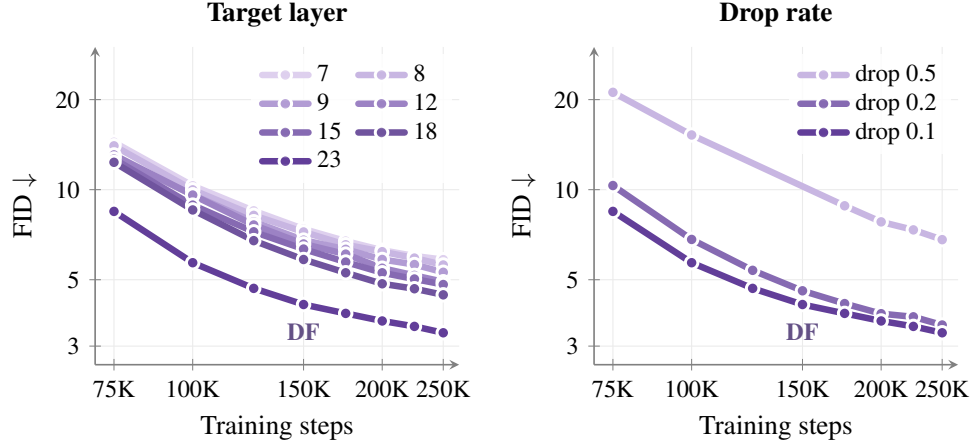

We now ablate the main design choices of our method. Unless stated otherwise, every ablation is run on \texttt{SiT-L} and reported in the unguided setting, so that all numbers are directly comparable.

\paragraph{Where should the pretrained representation live?} We sweep the block at which the pretrained representation prediction is attached and plot the resulting FID trajectories in \cref{fig:layer_index_abla} (left). Deeper is consistently better throughout training: the final block (\texttt{Layer 23}, zero-indexed) reaches $2.64$ against roughly $5.09$ for \texttt{Layer 7} at 250K steps. This is consistent with the branch requiring the abstract, high-level features that emerge only deep in the network, which are also the features that align with the \texttt{DINOv2} target. We attach the branch at the deepest block in all main experiments.

\paragraph{How often should we train on Pass~1 alone?} Pass~1 carries a specific burden: at sampling time it is the first step the model takes and its look-ahead must orient the trajectory before any real conditioning exists. Pass~1 is optimized alone with probability $p$. We sweep it in \cref{fig:layer_index_abla} (right). $p=0.1$ works best ($2.64$) compared to $p=0.2$ ($2.73$) and $p=0.5$ ($5.74$) at 250K steps. We use $p=0.1$ throughout.

\begin{figure*}[t]
    \centering
    \resizebox{\textwidth}{!}{\input{images/perturbation}}
    \caption{\textbf{Robustness to perturbing self-condition.} Gaussian noise with standard deviation $\sigma$ is added to the predicted \texttt{DINO} self-condition at a single sampling step $K$ (out of 50). \textbf{(Left)}~PSNR between perturbed and reference images, averaged over 256 images. \textbf{(Right)}~Top-1 accuracy of a linear probe on the final predicted \texttt{DINO} features. (\textit{\textbf{\underline{Note}}: Colour and marker size both encode the metric, with darker and larger being better}).}
    \label{fig:perturbation}
\end{figure*}

\paragraph{How does the representation perturbation affect the model?} We run two experiments to see the effect of perturbed self-condition measuring the impact on the generated image (Experiment~1) and the semantic content (Experiment~2). Each experiment perturbs the model's own \texttt{DINO} prediction (the self-condition) at a specific sampling step $K$ out of 50 by adding gaussian noise with zero mean and standard deviation $\sigma$, and then measures the effect on the final output (see \cref{fig:perturbation}).
 
\textit{Experiment 1: Effect on the generated image.} We generate 256 reference images with standard sampling. We then repeat sampling with the same seeds, but add noise to the self-condition at step $K$, and compute the PSNR between each perturbed image and its reference; a higher PSNR means the perturbation changed the image less (\cref{fig:perturbation}; left). For small perturbations ($\sigma \leq 0.25$), the final image stays very close to the reference (PSNR above 41\,dB). Larger perturbations change the image more. The effect is strongest at intermediate steps, with the lowest PSNR of 26.8\,dB at $K=10$ and $\sigma=2.0$. Perturbations at earlier steps are partly corrected and those at later steps barely change the image: errors do not accumulate, because the model predicts a new self-condition from the current image at every step.
 
\textit{Experiment 2: Effect on semantic content.} A lower PSNR shows that the image changed, but not whether its content is still correct. We therefore also test whether the final predicted \texttt{DINO} features still encode the right class. We first train a linear probing model on real \texttt{DINO} features (100K training features, 30 epochs, AdamW with lr $10e-3$ and cosine decay), which reaches 89.1\% top-1 accuracy on 5K held-out set. We then generate 2{,}000 \texttt{DINO} features with our model (2 per class), perturb the self-condition at step $K$ as in Experiment~1, and classify the final predicted features with this probe (\cref{fig:perturbation}; right). Class recovery stays close to the 89.1\% reference in every setting: even the strongest perturbations reduce it by at most 1.8 points (87.3\% at $\sigma=2.0$). Perturbations late in sampling ($K=35$) leave it unchanged, while at earlier steps the drop grows with $\sigma$. The perturbations therefore change image details but preserve semantic content.

\begin{table}[h]
\centering
\small
\setlength{\tabcolsep}{5pt}
\caption{Few-step sampling. FID\,$\downarrow$ of L-sized models on ImageNet $256^2$ (no CFG), sampled with the Euler ODE solver using different numbers of steps.}
\label{tab:few_step}
\begin{NiceTabular}{@{}l ccccc@{}}
\toprule
 & \multicolumn{5}{c}{\textbf{Sampling steps}} \\
\cmidrule(l){2-6}
\textbf{Model} & 10 & 15 & 20 & 25 & 50 \\
\midrule
\texttt{SiT-L/2} & 32.64 & 25.94 & 23.30 & 22.08 & 20.02 \\
\quad $\rightarrow$ \texttt{REPA} & 24.05 & 17.45 & 15.03 & 13.75 & 11.81 \\
\midrule
\rowcolor{dfcolor!7}\stripe{dfcolor}\quad $\rightarrow$ \textbf{\texttt{DF}}
    & \textbf{21.70} & \textbf{10.13} & \textbf{6.08} & \textbf{4.32} & \textbf{2.25} \\
\bottomrule
\end{NiceTabular}
\end{table}
 
\paragraph{How well does \mname{} perform in few step regime?} Since \mname{} refines its own \texttt{DINO} prediction along the sampling trajectory, a natural question is whether it needs many steps to benefit from it. We therefore sample \texttt{SiT-L/2}, \texttt{REPA} and \mname{} (all L-sized) with the Euler ODE solver using 10 to 50 steps (\cref{tab:few_step}). \mname{} achieves the best FID at every step count, so predicting the \texttt{DINO} representation does not hurt few-step generation. The advantage is smallest at 10 steps, where all models degrade heavily ($21.70$ vs.\ $24.05$ for \texttt{REPA}), and grows quickly with more steps: with only 15 steps, \mname{} ($10.13$) already outperforms \texttt{REPA} with 50 steps ($11.81$). As \mname{} uses a single forward pass per sampling step, this corresponds to $3.3\times$ fewer function evaluations for a better FID.

\section{Extended related work}
\label{app:sec:related_work}
\paragraph{Representation alignment in image generation.}
Diffusion models~\citep{sohl2015deep,ho2020denoising} and flow matching models~\citep{lipman2023flow} are trained with denoising objectives that predominantly encourage fine-grained, reconstruction-oriented internal representations rather than high-level semantic ones~\citep{yu2025representation}. This lack of semantic structure slows training convergence with state-of-the-art image generation models typically requiring over 500 training epochs to reach strong performance~\citetext{\abbrcite{DiT}{peebles2023scalable}; \abbrcite{SiT}{ma2024sit}; \abbrcite{JiT}{li2026back}; \abbrcite{$fv$-loss}{degeorge2026balancing}; \abbrcite{TREAD}{krause2025tread}}. Recent works~\citetext{\abbrcite{REPA}{yu2025representation}; \abbrcite{iREPA}{singh2026what}; \abbrcite{HASTE}{wang2025repa}; \abbrcite{REED}{wang2026learning}; \abbrcite{DDT}{wang2026ddt}; \abbrcite{PixelDiT}{yu2026pixeldit}; \abbrcite{PixelREPA}{shin2026representation}; \abbrcite{DeCo}{ma2026deco}; \abbrcite{PixelGen}{ma2026pixelgen}} have shown that explicitly aligning a model's internal features with external pretrained representations substantially accelerates convergence. A complementary line of work takes this further by performing flow matching directly in a pre-aligned representation latent space~\citetext{\abbrcite{RAE}{zheng2026diffusion}; \abbrcite{RAEv2}{singh2026improved}; \abbrcite{DINO-SAE}{chang2026dino}}, yielding additional convergence speedups.
Despite their promise, both approaches have their own limitations when applied to image generation. Feature alignment methods inject pretrained information only during training; at sampling time, the model operates without any explicit guidance, effectively discarding the alignment signal precisely when it is needed. Pre-aligned latent approaches sidestep this by embedding pretrained information into the generation space itself, but at the cost of training a new decoder and introducing non-trivial architectural modifications. A natural middle ground is to jointly denoise both pretrained representation and image, which we discuss next.
\paragraph{Co-denoising for image generation.}
Jointly denoising multiple modalities~\citep{bao2023one, xu2023versatile} offers an appealing alternative: it implicitly aligns pretrained representations while preserving them throughout the entire generation process, including at inference. Applied to image generation, this strategy accelerates training convergence, training both in a latent image feature space~\citetext{\abbrcite{ReDi}{kouzelis2026boosting}; \abbrcite{CoReDi}{kouzelis2026coevolving}; \abbrcite{REG}{wu2026representation}; \abbrcite{SFD}{pan2026semantics} \abbrcite{REGLUE}{petsangourakis2025reglue}} or directly in pixel space~\citetext{\abbrcite{LatentForcing}{baade2026latent}; \abbrcite{V-Co}{lin2026v}; \abbrcite{CoReDi}{kouzelis2026coevolving}}.
However, co-denoising introduces its own complexity: the optimal noise schedule for the pretrained representation is non-trivial to determine and must be carefully hand-crafted. For instance, \texttt{SFD} \citep{pan2026semantics} and \texttt{LatentForcing} \citep{baade2026latent} found that denoising the pretrained representation branch (e.g., \texttt{DINO} features) to a low noise level \emph{before} beginning image denoising yields the best convergence speed, a schedule that likely needs to be re-tuned for different modalities, making the approach brittle to apply in general.

The limitations above share a common structure: they all stem from either the absence of pretrained information during image generation, or the need for additional models and carefully designed schedules to introduce it. One principled way to address this is to condition the image generation model directly on pretrained representations, bypassing the need to denoise them from scratch ~\citetext{\abbrcite{RCG}{li2024return}; \abbrcite{PixelDiT2}{yu2026pixeldit2representationgroundedpixeldiffusion}; \abbrcite{WiT}{wang2026wit}}. Inspired by self-conditioning techniques~\citep{jabri2023scalable, chen2023analog}, we propose to condition the model on its \emph{own predicted representation} at each denoising step. Crucially, unlike prior self-conditioning works, we explicitly make these predicted self-conditions to align with an external pretrained feature space, encouraging the model to \emph{predict its own representation} in a meaningful, structured sense.  
This design simultaneously resolves the key failure modes of existing approaches: pretrained information is available throughout image generation without requiring a separate decoder, no architectural modifications, or handcrafted denoising schedules, while still achieving faster training convergence.

\newpage
\section{Additional Qualitative Results}
\label{app:sec:qualitative_results}

%
\begingroup
\tikzset{every picture/.append style={font=\sffamily}}

\def\qualroot{images/raw_jpg_pictures/appendix_qualitative_vis}          
\def\qshowslots{0}           
\def\dinotimes{0.0, 0.1, 0.2, 0.3, 0.4, 0.6, 0.8, 1.0}   
\def\dinoN{8}                
\def\dinosamples{2}          

\ifdefined\qunit\else\newlength{\qunit}\fi
\ifdefined\qgap \else\newlength{\qgap}\fi
\ifdefined\dunit\else\newlength{\dunit}\fi
\ifdefined\dlabw\else\newlength{\dlabw}\fi
\ifdefined\dgap \else\newlength{\dgap}\fi
\setlength{\qgap}{1.2pt}     
\setlength{\dlabw}{0.95cm}   
\setlength{\dgap}{1pt}

\def\qtile#1#2#3#4#5{%
  \node[anchor=north west, inner sep=0pt, outer sep=0pt]
    at ({#1\qunit + 0.5\qgap}, {-#2\qunit - 0.5\qgap})
    {\includegraphics[width=\dimexpr #3\qunit - \qgap\relax]{#4}};
  \ifnum\qshowslots=1
    \node[anchor=north west, fill=black, text=white, font=\sffamily\bfseries\tiny,
          inner sep=1.5pt, fill opacity=0.75, text opacity=1]
      at ({#1\qunit + 0.5\qgap}, {-#2\qunit - 0.5\qgap}) {#5};
  \fi
}

\def\mosaicEight#1#2{%
  \setlength{\qunit}{\dimexpr\linewidth/8\relax}%
  \begin{tikzpicture}
    \foreach [count=\i from 0] \c/\r/\s in {
        0/0/2, 4/0/2, 6/2/2, 2/4/2, 0/6/2, 6/6/2,
        2/0/1, 3/0/1, 6/0/1, 7/0/1, 2/1/1, 3/1/1,
        6/1/1, 7/1/1, 0/2/1, 1/2/1, 2/2/1, 3/2/1,
        4/2/1, 5/2/1, 0/3/1, 1/3/1, 2/3/1, 3/3/1,
        4/3/1, 5/3/1, 0/4/1, 1/4/1, 4/4/1, 5/4/1,
        6/4/1, 7/4/1, 0/5/1, 1/5/1, 4/5/1, 5/5/1,
        6/5/1, 7/5/1, 2/6/1, 3/6/1, 4/6/1, 5/6/1,
        2/7/1, 3/7/1, 4/7/1, 5/7/1}
    {
      \pgfmathtruncatemacro{\cc}{#1 ? 8-\c-\s : \c}
      \qtile{\cc}{\r}{\s}{#2/\i}{\i}
    }
  \end{tikzpicture}%
}

\def\mosaicSix#1#2{%
  \setlength{\qunit}{\dimexpr\linewidth/6\relax}%
  \begin{tikzpicture}
    \foreach [count=\i from 0] \c/\r/\s in {
        0/0/4,
        4/0/2, 4/2/2, 0/4/2, 4/4/2,
        2/4/1, 3/4/1, 2/5/1, 3/5/1}
    {
      \pgfmathtruncatemacro{\cc}{#1 ? 6-\c-\s : \c}
      \qtile{\cc}{\r}{\s}{#2/\i}{\i}
    }
  \end{tikzpicture}%
}

\def\dinosample#1#2{%
  \foreach [count=\k from 0] \t in \dinotimes {
    \node[anchor=north west, inner sep=0pt] at ({\dlabw + \k*\dunit}, {-#2*\dunit})
      {\includegraphics[width=\dimexpr\dunit-\dgap\relax]{#1/img_\k}};
    \node[anchor=north west, inner sep=0pt] at ({\dlabw + \k*\dunit}, {-(#2+1)*\dunit})
      {\includegraphics[width=\dimexpr\dunit-\dgap\relax]{#1/pca_\k}};
  }
  \node[rotate=90, font=\sffamily\tiny] at ({\dlabw-0.2cm}, {-(#2+0.5)*\dunit}) {Image};
  \node[rotate=90, font=\sffamily\tiny] at ({\dlabw-0.2cm}, {-(#2+1.5)*\dunit}) {DINOv2-B};
}
\def\dinoblock#1#2#3#4{%
  \foreach \s in {1,...,#3} {
    \pgfmathsetmacro{\yy}{#4 + 2.1*(\s-1)}
    \pgfmathtruncatemacro{\si}{\s-1}
    \dinosample{#1/s\si}{\yy}
  }
  \draw[line width=0.6pt, black!60] ({0.32cm}, {-#4*\dunit - 1pt})
    -- ({0.32cm}, {-(#4+2.1*#3-0.1)*\dunit + 1pt});
  \node[rotate=90, font=\sffamily\scriptsize\bfseries, inner sep=0pt]
    at ({0.15cm}, {-(#4+1.05*#3-0.05)*\dunit}) {#2};
}

\begin{figure}[h]
  \centering
  \mosaicEight{0}{\qualroot/qual/sit256_nocfg}
  \caption{Uncurated samples from DF-SiT-XL/2 on ImageNet $256{\times}256$, without guidance.}
  \label{fig:qual_sit256_nocfg}
\end{figure}

\begin{figure}[p]
  \centering
  \mosaicEight{1}{\qualroot/qual/sit256_cfg}
  \caption{Uncurated samples from DF-SiT-XL/2 on ImageNet $256{\times}256$, with cfg=4.0.}
  \label{fig:qual_sit256_cfg}
\end{figure}

\begin{figure}[p]
  \centering
  \mosaicSix{0}{\qualroot/qual/sit512_nocfg}
  \caption{Uncurated samples from DF-SiT-XL/2 on ImageNet $512{\times}512$ without guidance.}
  \label{fig:qual_sit512_nocfg}
\end{figure}

\begin{figure}[p]
  \centering
  \mosaicSix{1}{\qualroot/qual/sit512_cfg}
  \caption{Uncurated samples from DF-SiT-XL/2 on ImageNet $512{\times}512$, with cfg=4.0.}
  \label{fig:qual_sit512_cfg}
\end{figure}

\begin{figure}[p]
  \centering
  \mosaicEight{0}{\qualroot/qual/jit256_nocfg}
  \caption{Uncurated samples from DF-JiT-XL/16 on ImageNet $256{\times}256$ without guidance.}
  \label{fig:qual_jit256_nocfg}
\end{figure}

\begin{figure}[p]
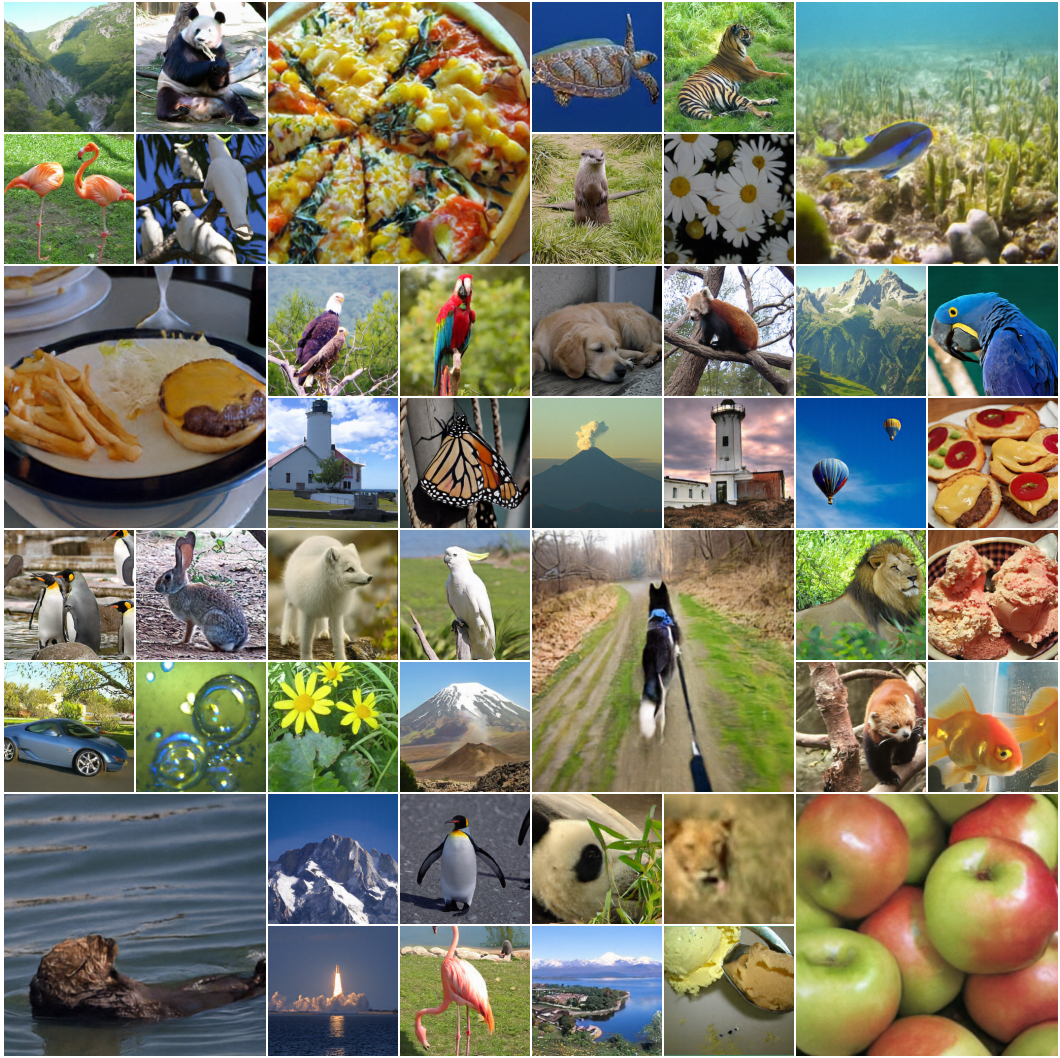

  \centering
  \mosaicEight{1}{\qualroot/qual/jit256_cfg}
  \caption{Uncurated samples from DF-JiT-XL/16 on ImageNet $256{\times}256$ with cfg=4.0.}
  \label{fig:qual_jit256_cfg}
\end{figure}

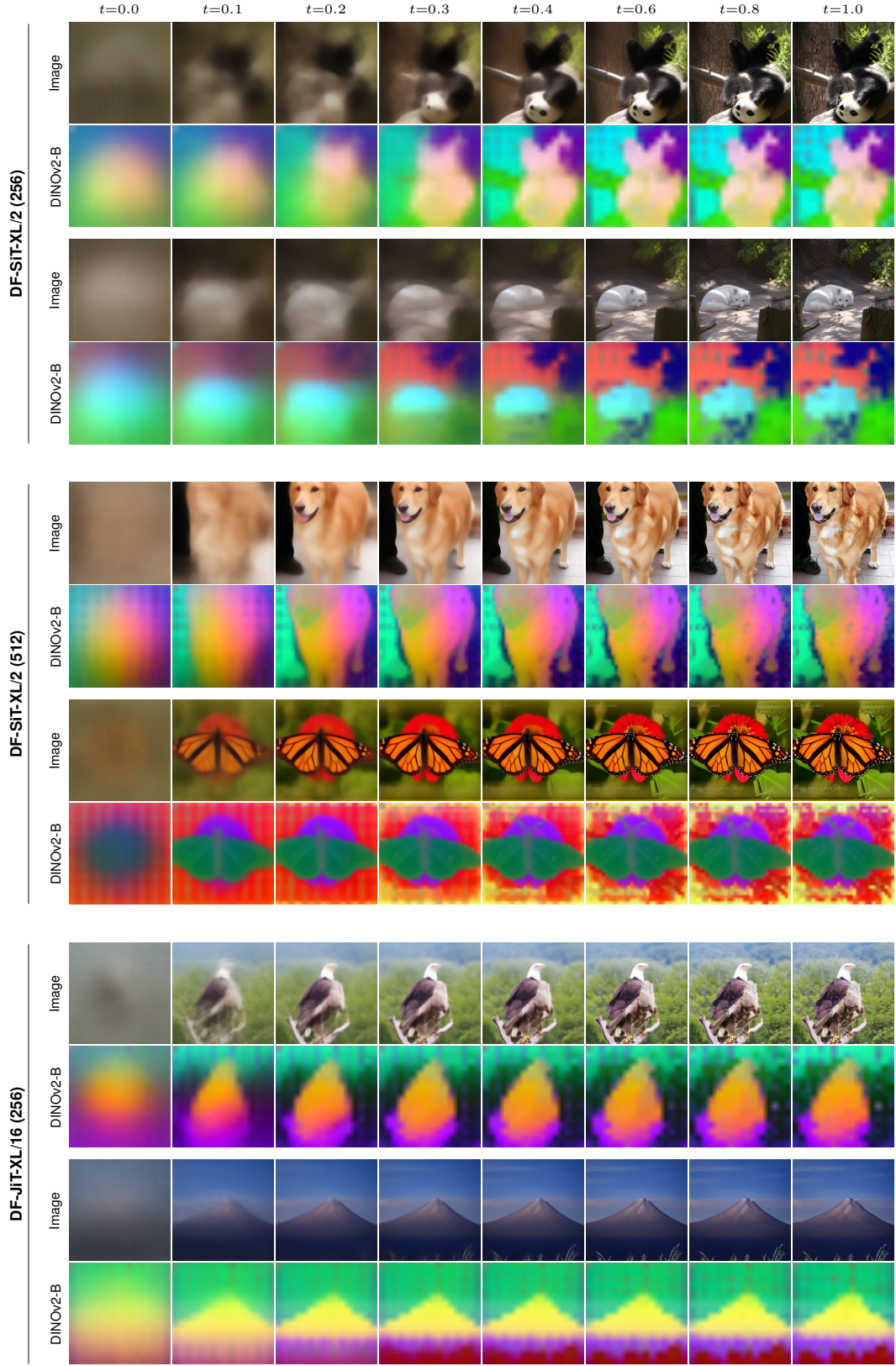
\begin{figure}[p]
  \centering
  \setlength{\dunit}{\dimexpr(\linewidth-\dlabw)/\dinoN\relax}%
  \begin{tikzpicture}
    \foreach [count=\k from 0] \t in \dinotimes {
      \node[anchor=south, font=\sffamily\tiny] at ({\dlabw + (\k+0.5)*\dunit - 0.5\dgap}, 0) {$t{=}\t$};
    }
    \pgfmathsetmacro{\bA}{0}
    \pgfmathsetmacro{\bB}{2.1*\dinosamples + 0.25}
    \pgfmathsetmacro{\bC}{4.2*\dinosamples + 0.5}
    \dinoblock{\qualroot/dino/sit256}{DF-SiT-XL/2 (256)}{\dinosamples}{\bA}
    \dinoblock{\qualroot/dino/sit512}{DF-SiT-XL/2 (512)}{\dinosamples}{\bB}
    \dinoblock{\qualroot/dino/jit256}{DF-JiT-XL/16 (256)}{\dinosamples}{\bC}
  \end{tikzpicture}
  \caption{Evolution of the image and the PCA of the predicted DINO features throughout
  sampling ($t{=}0$: noise, $t{=}1$: data), for DF-SiT-XL/2 at $256{\times}256$ and
  $512{\times}512$ and DF-JiT at $256{\times}256$. The PCA projection is fit once per sample
  on the final features and reused for all timesteps.}
  \label{fig:qual_dino_progression}
\end{figure}

\endgroup

\end{document}